\documentclass[letterpaper, 10 pt, conference]{ieeeconf}  

\IEEEoverridecommandlockouts                              

\let\labelindent\relax

\usepackage{macros}

\begin{document}

\title{\LARGE \bf Schur-MI: Fast Mutual Information for Robotic Information Gathering}

\author{Kalvik Jakkala, Jason O'Kane, and Srinivas Akella
\thanks{$^{1}$Kalvik Jakkala and Jason O'Kane are with Texas A\&M University
        {\tt\small \{kalvik, jokane\}@tamu.edu} 
$^{2}$Srinivas Akella is with University of North Carolina at Charlotte
        {\tt\small sakella@charlotte.edu}}%
}


\maketitle
\thispagestyle{empty}
\pagestyle{empty}

\maketitle

\begin{abstract}
Mutual information~(MI) is a principled and widely used objective for robotic information gathering~(RIG), providing strong theoretical guarantees for sensor placement~(SP) and informative path planning~(IPP). However, its high computational cost—dominated by repeated log-determinant evaluations—has limited its use in real-time planning. This paper presents \emph{Schur-MI}, a Gaussian process~(GP) MI formulation that (i) leverages the iterative structure of RIG to \emph{precompute} and reuse expensive intermediate quantities across planning steps, and (ii) uses a \emph{Schur-complement} factorization to avoid large determinant computations. Together, these methods reduce the per-evaluation cost of MI from $\mathcal{O}(|\mathcal{V}|^3)$ to $\mathcal{O}(|\mathcal{A}|^3)$, where $\mathcal{V}$ and $\mathcal{A}$ denote the candidate and selected sensing locations, respectively. Experiments on real-world bathymetry datasets show that Schur-MI achieves up to a $12.7\times$ speedup over the standard MI formulation. Field trials with an autonomous surface vehicle~(ASV) performing adaptive IPP further demonstrate the method’s practicality. By making MI computation tractable for online planning, Schur-MI helps bridge the gap between information-theoretic objectives and real-time robotic exploration. Our code is available at: \href{www.sgp-tools.com}{www.sgp-tools.com}
\end{abstract}

\IEEEpeerreviewmaketitle

\section{Introduction}

The ability of autonomous robots to efficiently map and model unknown environments is a cornerstone of modern robotics. This field, known as robotic information gathering~(RIG), underpins applications where human intervention is dangerous, inefficient, or impossible. Examples include monitoring the ocean floor and atmospheric phenomena, supporting search-and-rescue operations, and exploring planetary surfaces.

At the heart of RIG lie two fundamental challenges: sensor placement~(SP) and informative path planning~(IPP). The SP problem addresses the task of selecting a finite set of locations to deploy stationary sensors so as to maximize information about a spatial or spatiotemporal phenomenon. IPP generalizes this to a dynamic, sequential decision-making setting in which a mobile robot must choose a path—subject to constraints such as time or energy—to collect the most informative sequence of measurements.

\begin{figure}[ht!]
    \centering
    \includegraphics[width=0.95\linewidth]{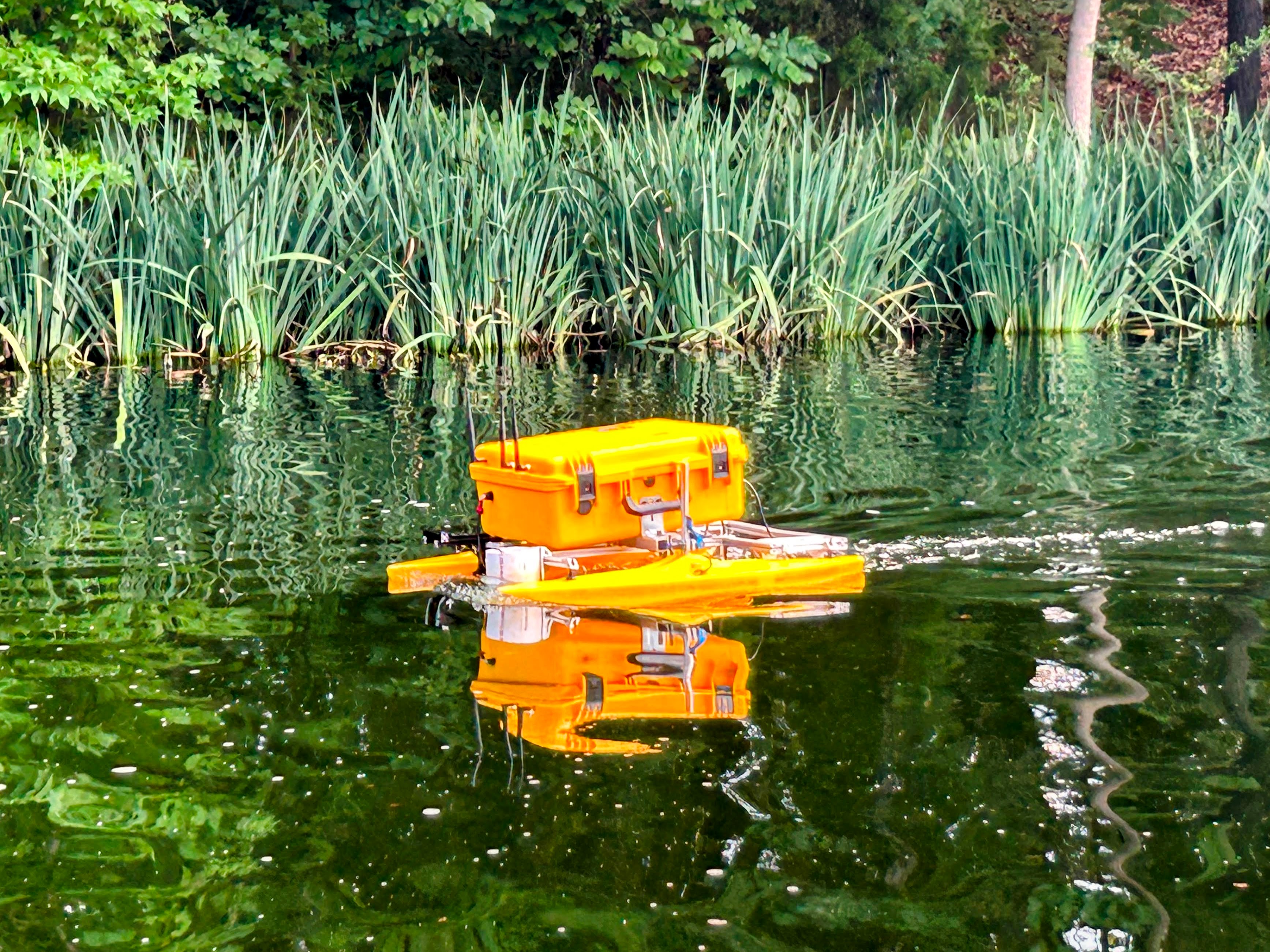}
    \caption{An autonomous surface vehicle mapping a lake during our field trials.} 
    \label{fig:cover}
\end{figure}

A dominant paradigm for both SP and IPP is to model the unknown environment with a Gaussian process~(GP), a non-parametric Bayesian model that provides both predictions and principled measures of uncertainty~\cite{RasmussenW05}. Within this framework, mutual information~(MI) has emerged as an effective objective function~\cite{CaseltonZ84, krauseSG08}. While simpler metrics like variance or entropy strictly target regions of maximum uncertainty, MI balances localized uncertainty with the predictive correlation of the surrounding space, making it a robust criterion for autonomous exploration.

Over the past two decades, approaches for optimizing MI have evolved significantly. Seminal work showed that maximizing the submodular MI objective with a greedy algorithm over a discrete set of candidate locations yields \emph{near-optimal} solutions~\cite{krauseSG08}. To extend beyond discrete search, subsequent work has focused on continuous-space optimization using derivative-free, black-box methods such as genetic algorithms~\cite{HitzGGPS17} and Bayesian optimization~\cite{FrancisOMR19, LinCWT19}, enabling direct optimization of complex path parameterizations.

Despite its strengths, MI remains underused in practice due to a fundamental \emph{computational bottleneck}: each MI evaluation requires computing matrix determinants at a cost of $\mathcal{O}(m^3)$, where $m$ is the number of candidate sensing locations in the environment. Because modern planners—from greedy selection to advanced adaptive frameworks—require numerous objective evaluations, this cubic cost often forces practitioners to abandon MI in favor of computationally cheaper but less informative proxies~\cite{OttKB24, JakkalaA25}. Thus, the gap between MI’s theoretical appeal and its practical feasibility has persisted for decades.

This paper addresses this long-standing challenge by first identifying two key computational bottlenecks in the standard GP MI criterion. We then introduce precomputation to amortize terms that are independent of the selected sensing set, mitigating the first bottleneck. To address the second bottleneck, we propose an alternate RIG formulation that exposes additional reusable structure; however, this reformulation can both introduce a degeneracy and, in certain regimes, increase computational cost. We resolve these issues by deriving Schur-MI, which removes the degeneracy via a perturbed surrogate candidate set and substantially reduces computation cost by leveraging a Schur-complement factorization with precomputation, making MI-based planning feasible for real-time, adaptive, and large-scale RIG.
\section{Related Work}

Our work, which aims to accelerate the computation of mutual information~(MI), is framed within three key areas of research: the formulation of information-theoretic objectives, the design of planners that optimize them, and the need for scalable computation.

\subsection{Information-Theoretic Objectives for RIG}

The effectiveness of an informative path is determined by the objective function it seeks to optimize. While early approaches relied on simple geometric criteria like coverage~\cite{Ramsden09, SchwagerVPRT17}, modern methods leverage probabilistic models of the environment to guide planning~\cite{krauseSG08}. Gaussian processes~(GPs)~\cite{RasmussenW05} have become the predominant model, giving rise to several information-theoretic metrics.

A common and computationally efficient metric is the predictive variance (or entropy) at unsensed locations. The strategy of greedily sampling points of highest variance is intuitive but can lead to suboptimal placements, often clustering sensors along the boundaries of the search space~\cite{krauseSG08}. Other classical criteria from the field of optimal experimental design, such as A-, D-, and B-optimality~\cite{OttKB24}, have also been adapted but suffer from similar limitations.

The MI criterion, proposed by Caselton and Zidek~\cite{CaseltonZ84}, stands out as a more robust metric. By quantifying the reduction in uncertainty over all \textit{unobserved} locations, MI naturally encourages placements that are centrally located and highly informative.

Krause et al.~\cite{krauseSG08} demonstrated that although
maximizing MI is NP-complete, its submodularity enables a greedy algorithm to achieve a \emph{near-optimal solution with a $(1 - 1/e)$ approximation guarantee}. Furthermore they proposed an efficient approximation that sparsifies the covariance matrix to reduce inversion costs, thereby lowering the overall computational complexity from $\mathcal{O}(sm^4)$ to $\mathcal{O}(sm)$, where $s$ and $m$ denote the numbers of sensing and candidate locations, respectively. However, this sparsification approach sacrifices critical long-range correlations captured by non-stationary kernels, thus limiting its applicability to stationary settings.

\subsection{Optimization and Planning Algorithms}

Given an information objective, a planner’s role is to find a path that optimizes the objective. Key methods are summarized below:

\textit{Discrete-Space Greedy Optimization:} The foundational method, enabled by MI’s submodularity, is a greedy algorithm that iteratively adds the most informative location from a discrete set of candidates. While this approach is straightforward and provides near-optimal guarantees, it remains restricted to a predefined set of candidate locations~\cite{krauseSG08, BinneyKS13}.

\textit{Continuous-Space Optimization:} To plan paths in continuous spaces, researchers have employed derivative-free, black-box optimization techniques like genetic algorithms~\cite{HitzGGPS17} and Bayesian optimization~\cite{FrancisOMR19, LinCWT19}. These methods can handle complex path parameterizations but treat the MI objective as a black box, requiring numerous expensive evaluations that worsen the computational bottleneck.

\textit{Adaptive Planning:} More advanced planners address the adaptive nature of IPP, where decisions are made online as new data is collected~\cite{OttBK23, OttKB24}. These methods often frame the problem as a partially observable Markov decision process~(POMDP) and use techniques like Monte Carlo tree search~(MCTS). Notably, these planners often abandon the MI objective for computationally cheaper proxies like variance reduction to maintain tractability.

\textit{Learning-Based Approaches:} A recent trend involves using machine learning, particularly Reinforcement Learning (RL) and Imitation Learning (IL), to learn complex, reactive planning policies~\cite{ManjannaD17, RuckinJP23}. While powerful, these methods do not eliminate the underlying issue; if the policy is trained using an MI-based reward signal, the training process itself will require a massive number of expensive MI calculations.

\subsection{Scalability and Efficiency in RIG}

The computational burden of GP-based planning with the MI objective is a major theme in literature. As robotic applications grow in scale and complexity, so does the need for efficient methods. Prior research has primarily addressed this bottleneck through the following approaches:

\textit{Approximating the Objective:} One common strategy is to replace or approximate the MI objective~\cite{AsgharivaskasiKA22, JakkalaA25b}. Some methods use simpler proxies like variance reduction or develop lower-dimensional projections to create a bound on MI. Others create differentiable approximations of MI to enable faster, gradient-based optimization~\cite{OttKB24, JakkalaA25}. Notably, Ott et al.~\cite{OttBK23} proposed approximating MI by replacing determinant operations with trace operations.

\textit{Improving the Model:} Another approach focuses on the underlying GP model itself~\cite{TajnafoiAMVMPG21, ChenKL22, ChenLK24}. This includes using more efficient model variants like variational or sparse Gaussian processes to handle large datasets or developing non-stationary kernels for more accurate uncertainty estimates. These methods improve the model from which MI is computed but do not accelerate the MI calculation itself.

Our work addresses a critical gap left by these complementary efforts. While planners have grown more powerful and models more expressive, the MI objective has remained computationally prohibitive. Instead of relying on approximations, we tackle this bottleneck directly. Leveraging precomputation and the Schur complement, we accelerate MI computation, enabling advanced, large-scale, and adaptive planning frameworks to fully realize the potential of state-of-the-art robotic information gathering.
\section{Preliminaries: Gaussian processes}

To formally model a phenomenon and quantify information, we use Gaussian processes~(GPs)~\cite{RasmussenW05}, a powerful non-parametric tool from Bayesian statistics for modeling correlated data.

A Gaussian process is a collection of random variables, any finite number of which have a joint Gaussian distribution. Despite their name, GPs are not limited to modeling phenomena that are themselves Gaussian distributed. Rather, a GP defines a distribution over functions, making it a flexible tool for a wide range of regression problems.

The method places a GP prior over an unknown function:

\begin{equation}
\begin{aligned}
    f(\mathbf{x}) \sim \mathcal{GP}(m(\mathbf{x}), k(\mathbf{x}, \mathbf{x}')) \,.
\end{aligned}
\end{equation}

This process is completely specified by a mean function $m(\mathbf{x})$ and a covariance function (or kernel) $ k(\mathbf{x}, \mathbf{x}')$. The mean function, often assumed to be zero after data normalization, represents the expected value of the function at location $\mathbf{x}$. The kernel function defines the covariance between the function values at two locations, $\mathbf{x}$ and $\mathbf{x}'$:

\begin{equation}
\begin{aligned}
    k(\mathbf{x}, \mathbf{x}') = \text{cov}(f(\mathbf{x}), f(\mathbf{x}')) \,.
\end{aligned}
\end{equation}

The kernel encodes our prior beliefs about the properties of the function, such as its smoothness and length-scale. A common choice is the squared exponential (or radial basis function (RBF)) kernel:

\begin{equation}
\begin{aligned}
    k(\mathbf{x}, \mathbf{x}') = \sigma^2_f \text{exp}\left( -\frac{||\mathbf{x}-\mathbf{x}'||^2}{2l^2} \right) \,,
\end{aligned}
\end{equation}

\noindent where the signal variance $\sigma^2_f$ controls the overall variance of the function, and the length-scale $l$ determines how quickly the correlation between points decays with distance.

Given a set of $n$ noisy observations $\mathcal{D} = \{\mathbf{X}, \mathbf{y}\} =  \{(\mathbf{x}_i, y_i), i = 1,...,n\}$, where $y_i = f(\mathbf{x}_i) + \epsilon_i$ and the noise $\epsilon_i$ is assumed to be i.i.d. Gaussian with variance $\sigma^2_n$, we can compute the posterior predictive distribution for a new test point $\mathbf{x}_*$. This posterior is also a GP, $p(f_*|\mathbf{X}, \mathbf{y}, \mathbf{x}_*) = \mathcal{GP}(\mu_*, \sigma^2_*)$, with mean $\mu_*$ and variance $\sigma^2_*$ given by:

\begin{equation}
\begin{aligned}
\label{eq:gp_pred}
\mu_* &= \mathbf{K}_{*n}(\mathbf{K}_{nn} + \sigma^2_n \mathbf{I})^{-1}\mathbf{y}\,, \\
\sigma^2_* &= \mathbf{K}_{**} - \mathbf{K}_{*n}(\mathbf{K}_{nn} + \sigma_n^2 \mathbf{I})^{-1}\mathbf{K}_{n*}\,,
\end{aligned}
\end{equation}

\noindent where $\mathbf{K} = k(\cdot, \cdot)$ are the covariance matrices. The covariance matrix subscripts indicate the variables used to compute it; $*$ for the test input $\mathbf{x}_*$ and $n$ for the training inputs $\mathbf{X}$. 
\section{Problem Statement}
Robotic information gathering (RIG) concerns the problem of selecting informative sampling locations to efficiently estimate an unknown spatial or spatiotemporal field, $\mathcal{X} \subset \mathbb{R}^d$. For the purposes of estimation and decision-making, the environment is discretized into a set of $m$ evaluation points, $\mathcal{V} = \{x_1, \ldots, x_m\} \subset \mathcal{X}$. The goal is to infer an underlying field $y : \mathcal{V} \to \mathbb{R}^p$, representing environmental quantities such as ocean temperature, salinity, or terrain elevation.

Our objective is to select a subset of $s$ sampling locations, $\mathcal{A} = \{x_i \in \mathcal{V} \mid i = 1, \ldots, s\}$, with $s \ll m$, at which noisy measurements are obtained as $y_i = y(x_i) + \epsilon_i$, where $\epsilon_i \sim \mathcal{N}(0, \sigma^2_\epsilon)$ models sensor noise. The collected data must maximize the expected information about $y(\cdot)$ over $\mathcal{V}$ according to an information metric $\mathbb{I}(\cdot)$.

This work focuses on \emph{mutual information}~(MI) as the optimization criterion, owing to its strong theoretical foundations and near-optimal approximation guarantees. The MI-based RIG problem is typically formulated as selecting a subset of sensing locations $\mathcal{A}$ that maximizes the MI between the sampled and unsampled regions:
\begin{equation}
\label{eq:mi_rig}
    \mathcal{A}^* = \argmax_{\mathcal{A} \subset \mathcal{V}} \ \mathbb{I}(\mathcal{A}; \mathcal{V} \setminus \mathcal{A}) \,,
\end{equation}
The MI between these two sets of random variables is defined as follows, where $H(\cdot)$ denotes Shannon entropy~\cite{Bishop06}:
\begin{equation}
\label{eq:mi_set}
\begin{aligned}
\mathbb{I}(\mathcal{A}; \mathcal{V} \setminus \mathcal{A}) 
&= H(\mathcal{A}) - H(\mathcal{A} \mid \mathcal{V} \setminus \mathcal{A}) \\
&= H(\mathcal{A}) + H(\mathcal{V} \setminus \mathcal{A}) - H(\mathcal{V}) \,.
\end{aligned}
\end{equation}

We model the underlying phenomenon using a Gaussian process~(GP), leveraging its covariance structure for closed-form MI computation. Optimizing (\ref{eq:mi_rig}) for MI-based RIG requires evaluating MI across many candidate sets, $\mathcal{A}$. However, evaluating MI under a GP is computationally demanding, requiring $\mathcal{O}(m^3)$ operations for the necessary matrix inversions and determinant evaluations. This computational bottleneck significantly limits the scalability of MI-based methods for real-time and large-scale planning. Therefore, the central problem addressed in this work is reducing the computational cost of MI evaluations for RIG.

To overcome this challenge, we propose an efficient formulation of MI for RIG that preserves the theoretical rigor of the original objective while substantially reducing computational complexity. This approach enables principled, MI-driven decision-making for real-world, time-critical robotic exploration and sensing.

\section{Method}

Our goal is to select sensing locations that maximize mutual information~(MI) under a Gaussian process~(GP) model, while keeping the computation fast enough for real-time robotic information gathering~(RIG). We first introduce the standard GP MI criterion and identify its dominant computation cost bottlenecks. Next, we show how simple precomputation amortizes repeated costs and motivates an alternate RIG objective that exposes additional reusable structure. Finally, we derive an equivalent Schur-complement MI formulation which, combined with precomputation and a nondegenerate surrogate candidate set, yields an efficient objective suitable for real-time RIG.

\subsection{Standard Mutual Information Criterion}

Under a GP prior, the entropy of the latent variables~\cite{Bishop06} evaluated at a set of inputs $\mathcal{A}$ with covariance $\mathbf{K}_{\mathcal{A}\mathcal{A}}$ is

\begin{equation}
H(\mathcal{A}) = \frac{s}{2}\ln(2\pi e) + \frac{1}{2}\ln \det(\mathbf{K}_{\mathcal{A}\mathcal{A}}) ,
\end{equation}
where $s = |\mathcal{A}|$ and $\det(\cdot)$ denotes the matrix determinant. For clarity of exposition, we omit the observation noise variance $\sigma_n^2$ (equivalently, one may replace $\mathbf{K}$ by $\mathbf{K}+\sigma^2_\epsilon\mathbf{I}$).

Substituting this entropy into the set-based MI definition~(\ref{eq:mi_set}) gives:

\begin{equation}
\label{eq:mi_base}
\begin{aligned}
\mathbb{I}&(\mathcal{A}; \mathcal{V} \setminus \mathcal{A}) 
= H(\mathcal{A}) + H(\mathcal{V} \setminus \mathcal{A}) - H(\mathcal{V}) \\
&= \cancel{\frac{s}{2}\ln(2\pi e)} + \frac{1}{2}\ln\det(\mathbf{K}_{\mathcal{A}\mathcal{A}}) \\
&\ \ \ + \cancel{\frac{r}{2}\ln(2\pi e)} 
    + \frac{1}{2}\ln\det(\mathbf{K}_{(\mathcal{V}\setminus\mathcal{A})(\mathcal{V}\setminus\mathcal{A})}) \\
&\ \ \ - \cancel{\frac{m}{2}\ln(2\pi e)} - \frac{1}{2}\ln\det(\mathbf{K}_{\mathcal{V}\mathcal{V}}) \\
&= \frac{1}{2}\ln\left[
    \frac{
        \det(\mathbf{K}_{\mathcal{A}\mathcal{A}})
        \cdot
        \det(\mathbf{K}_{(\mathcal{V}\setminus\mathcal{A})(\mathcal{V}\setminus\mathcal{A})})
    }{
        \det(\mathbf{K}_{\mathcal{V}\mathcal{V}})
    }
\right] \,.
\end{aligned}
\end{equation}

Here, $m = |\mathcal{V}|$ and $r = |\mathcal{V}\setminus\mathcal{A}| = m-s$. Evaluating (\ref{eq:mi_base}) is dominated by computing $\det(\mathbf{K}_{\mathcal{V}\mathcal{V}})$ at cost $\mathcal{O}(m^3)$, followed by $\det\!\big(\mathbf{K}_{(\mathcal{V}\setminus\mathcal{A})(\mathcal{V}\setminus\mathcal{A})}\big)$ at cost $\mathcal{O}(r^3)$. In typical RIG settings $s \ll r < m$, so repeated evaluation of these log-determinants quickly becomes the computational bottleneck. We refer to (\ref{eq:mi_base}) as the \emph{Standard-MI} formulation.

\subsection{Precomputation and an Alternate MI-RIG Objective}

In MI-based RIG, the objective is evaluated repeatedly for many candidate sensing sets $\mathcal{A}$. This motivates \emph{amortizing} computations that do not depend on $\mathcal{A}$. In particular, in Standard-MI~(\ref{eq:mi_base}) the term $\det(\mathbf{K}_{\mathcal{V}\mathcal{V}})$ depends only on the fixed candidate set $\mathcal{V}$ and can therefore be computed once and reused. With this precomputation, the per-evaluation cost drops from $\mathcal{O}(m^3)$ to $\mathcal{O}(r^3)$, and the dominant remaining computation becomes
$\det\!\big(\mathbf{K}_{(\mathcal{V}\setminus\mathcal{A})(\mathcal{V}\setminus\mathcal{A})}\big)$.

However, under the standard RIG objective (\ref{eq:mi_rig}), the selected set $\mathcal{A}$ appears in both arguments of the MI term, so $\mathcal{V}\setminus\mathcal{A}$ changes across evaluations and the expensive determinant $\det\!\big(\mathbf{K}_{(\mathcal{V}\setminus\mathcal{A})(\mathcal{V}\setminus\mathcal{A})}\big)$ cannot be reused. To expose additional reusable structure, we instead consider the alternate objective

\begin{equation}
    \mathcal{A}^* = \argmax_{\{\mathcal{A} \subset \mathcal{V} \}} \  \mathbb{I}(\mathcal{A}; \mathcal{V}) \,,
\end{equation}
for which MI can be written as

\begin{equation}
\label{eq:mi_v2}
\begin{aligned}
\mathbb{I}&(\mathcal{A}; \mathcal{V}) 
= H(\mathcal{A}) + H(\mathcal{V}) - H(\mathcal{V} \cup \mathcal{A}) \\
&= \frac{1}{2}\ln\left[
    \frac{
        \det(\mathbf{K}_{\mathcal{A}\mathcal{A}})
        \cdot
        \det(\mathbf{K}_{\mathcal{V}\mathcal{V}})
    }{
        \det(\mathbf{K}_{(\mathcal{A \cup V})(\mathcal{A \cup V})})
    }
\right] \,.
\end{aligned}
\end{equation}

This removes the problematic term
$\det\!\big(\mathbf{K}_{(\mathcal{V}\setminus\mathcal{A})(\mathcal{V}\setminus\mathcal{A})}\big)$, but introduces the determinant over the union set $\det(\mathbf{K}_{(\mathcal{A \cup V})(\mathcal{A \cup V})})$. Also, if $\mathcal{A}\subset\mathcal{V}$, then $\mathcal{A}\cup\mathcal{V}=\mathcal{V}$ and the expression collapses to

\begin{equation}
\label{eq:mi_v3}
\begin{aligned}
\mathbb{I} (\mathcal{A}; \mathcal{V}) &= \frac{1}{2}\ln\left[
    \frac{
        \det(\mathbf{K}_{\mathcal{A}\mathcal{A}})
        \cdot
        \det(\mathbf{K}_{\mathcal{V}\mathcal{V}})
    }{
        \det(\mathbf{K}_{(\mathcal{A \cup V})(\mathcal{A \cup V})})
    }
\right] \\
&= \frac{1}{2}\ln\left[
    \frac{
        \det(\mathbf{K}_{\mathcal{A}\mathcal{A}})
        \cdot
        \cancel{\det(\mathbf{K}_{\mathcal{V}\mathcal{V}})}
    }{
        \cancel{\det(\mathbf{K}_{\mathcal{V}\mathcal{V}})}
    }
\right] \\
&= \frac{1}{2}\ln\left[
        \det(\mathbf{K}_{\mathcal{A}\mathcal{A}})
\right] \,,
\end{aligned}
\end{equation}
i.e., the objective degenerates to the entropy of $\mathcal{A}$.

While this loses the MI coupling between the selected set $\mathcal{A}$ and the unsampled locations $\mathcal{V \setminus A}$, the next subsection shows how to recover the exact MI using a nondegenerate surrogate construction together with a Schur-complement factorization, enabling further computational savings.

\subsection{Nondegenerate Schur-Complement-Based MI: Schur-MI}

We now resolve the degeneracy that occurs when $\mathcal{A}\subset\mathcal{V}$. Our key idea is to break the exact set inclusion by introducing a \emph{noisy surrogate} of the candidate set:
\[
\mathcal{G} \triangleq \mathcal{V}+\epsilon,\qquad \epsilon \sim \mathcal{N}(\mathbf{0},\sigma^2\mathbf{I}),
\]
with $\sigma^2$ chosen as a small fraction of the unit distance so that $\mathcal{G}$ remains a near-copy of $\mathcal{V}$ but does not coincide with it exactly. We then solve

\begin{equation}
    \mathcal{A}^* = \argmax_{\mathcal{A}\subset \mathcal{G}} \ \mathbb{I}(\mathcal{A}; \mathcal{V}) \,.
\end{equation}

Unlike the degenerate case, this objective does not collapse to the entropy of $\mathcal{A}$; it preserves the full MI coupling between the selected set $\mathcal{A}$ and the candidate set $\mathcal{V}$.

However, naively evaluating $\mathbb{I}(\mathcal{A};\mathcal{V})$ using~(\ref{eq:mi_v2}) requires computing determinants over the union set $\mathcal{A \cup V}$. When $\mathcal{A}\not\subset\mathcal{V}$, this operation incurs a computational complexity of $\mathcal{O}((s+m)^3)$, which scales more poorly than the Standard-MI formulation in~(\ref{eq:mi_base}). We circumvent this computational burden by leveraging the \emph{Schur complement}. Specifically, the joint covariance matrix over $\mathcal{A} \cup \mathcal{V}$ can be partitioned into the following block form:

\begin{equation}
\mathbf{K}_{(\mathcal{A \cup V})(\mathcal{A \cup V})} = 
\begin{bmatrix}
  \mathbf{K}_{\mathcal{A}\mathcal{A}} & \mathbf{K}_{\mathcal{A}\mathcal{V}} \\
  \mathbf{K}_{\mathcal{V}\mathcal{A}} & \mathbf{K}_{\mathcal{V}\mathcal{V}} \\
\end{bmatrix} \,,
\end{equation}
the determinant of this block matrix can be expressed via the Schur-complement~\cite{Bishop06} as:

\begin{equation}
\begin{aligned}
\label{eq:det_kvv}
\det(\mathbf{K}_{(\mathcal{A \cup V})(\mathcal{A \cup V})}) &= \det(\mathbf{K}_{\mathcal{V}\mathcal{V}}) \\
&\ \ \ \ \ \cdot \det\left(\mathbf{K}_{\mathcal{A}\mathcal{A}} 
   - \mathbf{K}_{\mathcal{A}\mathcal{V}} 
     \mathbf{K}_{\mathcal{V}\mathcal{V}}^{-1} 
     \mathbf{K}_{\mathcal{V}\mathcal{A}}\right) \\
&= \det(\mathbf{K}_{\mathcal{V}\mathcal{V}}) 
   \cdot \det(\mathbf{K}_{\mathcal{A}|\mathcal{V}}) \,,
\end{aligned}
\end{equation}
where $\mathbf{K}_{\mathcal{A}\mid\mathcal{V}}$ is the conditional covariance of $\mathcal{A}$ given $\mathcal{V}$. Substituting~(\ref{eq:det_kvv}) into~(\ref{eq:mi_v2}) yields the \emph{Schur-MI} objective:

\begin{equation}
\label{eq:mi_schur}
\begin{aligned}
\mathbb{I}(\mathcal{A}; \mathcal{V}) 
&= \frac{1}{2}\ln\left[
    \frac{
        \det(\mathbf{K}_{\mathcal{A}\mathcal{A}})
        \cdot
        \det(\mathbf{K}_{\mathcal{V}\mathcal{V}})
    }{
        \det(\mathbf{K}_{(\mathcal{A \cup V})(\mathcal{A \cup V})})
    }
\right] \\
&= \frac{1}{2}\ln\left[
\frac{\det(\mathbf{K}_{\mathcal{A}\mathcal{A}})\cdot
      \cancel{\det(\mathbf{K}_{\mathcal{V}\mathcal{V}})}}
     {\cancel{\det(\mathbf{K}_{\mathcal{V}\mathcal{V}})}
      \cdot \det(\mathbf{K}_{\mathcal{A}\,|\,\mathcal{V}})}
\right] \\
&= \frac{1}{2}\ln\left[
\frac{\det(\mathbf{K}_{\mathcal{A}\mathcal{A}})}
     {\det\left(\mathbf{K}_{\mathcal{A}\mathcal{A}} 
   - \mathbf{K}_{\mathcal{A}\mathcal{V}} 
     \mathbf{K}_{\mathcal{V}\mathcal{V}}^{-1} 
     \mathbf{K}_{\mathcal{V}\mathcal{A}}\right)}
\right] \,.
\end{aligned}
\end{equation}

Schur-MI is mathematically equivalent to Standard-MI~(\ref{eq:mi_base}), but exposes substantial structure for efficient evaluation. In particular, $\mathbf{K}_{\mathcal{V}\mathcal{V}}^{-1}$ can be computed once and reused across all candidate sets. With this precomputation, evaluating (\ref{eq:mi_schur}) requires only forming $\mathbf{K}_{\mathcal{A}\mathcal{V}}$ and computing determinants of $s\times s$ matrices, resulting in an effective cost of $\mathcal{O}(s^3)$ per MI evaluation.

Crucially, this formulation preserves the informativeness of the original MI-RIG objective $\mathbb{I}(\mathcal{A};\mathcal{V}\setminus\mathcal{A})$: it captures information about the full environment while avoiding the degeneracy that arises when $\mathcal{A}\subset\mathcal{V}$. Empirically, our experiments (Section~\ref{exp:sb2}) show that, under greedy selection with a stationary GP kernel, the proposed RIG objective—combined with the noisy-surrogate candidate-set construction, Schur-MI, and precomputation—achieves the same Standard-MI value for the selected sensing locations as the original RIG formulation (\ref{eq:mi_rig}) using Standard-MI.

\subsection{Reference Algorithm: Greedy Maximization of Schur-MI}
Algorithm~\ref{alg:SchurMI} summarizes the greedy approach for maximizing the proposed MI objective. Greedy MI maximization admits a \emph{near-optimal approximation guarantee} due to submodularity~\cite{krauseSG08}.

We present greedy selection as a reference rather than a requirement: the proposed MI-RIG formulation can be optimized with a range of alternatives, including Bayesian optimization~\cite{FrancisOMR19} and genetic algorithms~\cite{HitzGGPS17}. Moreover, it readily accommodates practical constraints such as distance budgets and motion feasibility, consistent with prior work~\cite{OttBK23, OttKB24}.

\begin{algorithm}[h]
\DontPrintSemicolon
\KwIn{Candidate locations $\mathcal{V}$, number of sensing locations $s$, kernel $k(\cdot,\cdot)$, noise $\sigma^2$.}
\KwOut{Solution locations $\mathcal{A}^*$.}

\tcp{Noisy surrogate set}
Sample $\epsilon \sim \mathcal{N}(\mathbf{0},\sigma^2\mathbf{I})$ and set $\mathcal{G}\leftarrow \mathcal{V}+\epsilon$\;

\tcp{Precompute inverse}
$\mathbf{K}_{\mathcal{V}\mathcal{V}} \leftarrow k(\mathcal{V},\mathcal{V}); \mathbf{K}_{\text{const}} \leftarrow \mathbf{K}_{\mathcal{V}\mathcal{V}}^{-1}$\;

\tcp{Initialize}
$\mathcal{A}^*\leftarrow \emptyset$; \ $\delta_{\text{base}}\leftarrow 0$\;

\For(\tcp*[f]{Greedy selection}){$j\leftarrow 1$ \KwTo $s$}{
  \ForEach(\tcp*[f]{Marginal gain}){$a\in \mathcal{G}$}{
    $\mathcal{A}\leftarrow \mathcal{A}^*\cup\{a\}$\;
    $\delta(a) \leftarrow \frac{1}{2}\ln\!\left[
        \frac{\det(\mathbf{K}_{\mathcal{A}\mathcal{A}})}
             {\det\big(\mathbf{K}_{\mathcal{A}\mathcal{A}} -
                  \mathbf{K}_{\mathcal{A}\mathcal{V}}\mathbf{K}_{\text{const}}
                  \mathbf{K}_{\mathcal{A}\mathcal{V}}^\top\big)}
      \right]-\delta_{\text{base}}$\;
  }

  \tcp{Select and update solution set}
  $a^* \leftarrow \arg\max_{a\in \mathcal{G}} \delta(a)$\;
  $\mathcal{A}^*\leftarrow \mathcal{A}^*\cup\{a^*\}$\;

  \tcp{Update baseline}
  $\delta_{\text{base}} \leftarrow \frac{1}{2}\ln\!\left[
      \frac{\det(\mathbf{K}_{\mathcal{A}^*\mathcal{A}^*})}
           {\det\!\big(\mathbf{K}_{\mathcal{A}^*\mathcal{A}^*} -
                \mathbf{K}_{\mathcal{A}^*\mathcal{V}}\mathbf{K}_{\text{const}}
                \mathbf{K}_{\mathcal{A}^*\mathcal{V}}^\top\big)}
    \right]$\;
}

\Return{$\mathcal{A}^*$}\;
\caption{Greedy sensor placement via Schur-MI with precomputation.}
\label{alg:SchurMI}
\end{algorithm}
\section{Experiments}

This section details the experiments used to evaluate the effectiveness and computational efficiency of Schur-MI.

\begin{figure*}[!ht]
   \centering
    \begin{subfigure}{0.24\textwidth}
        \includegraphics[width=\textwidth]{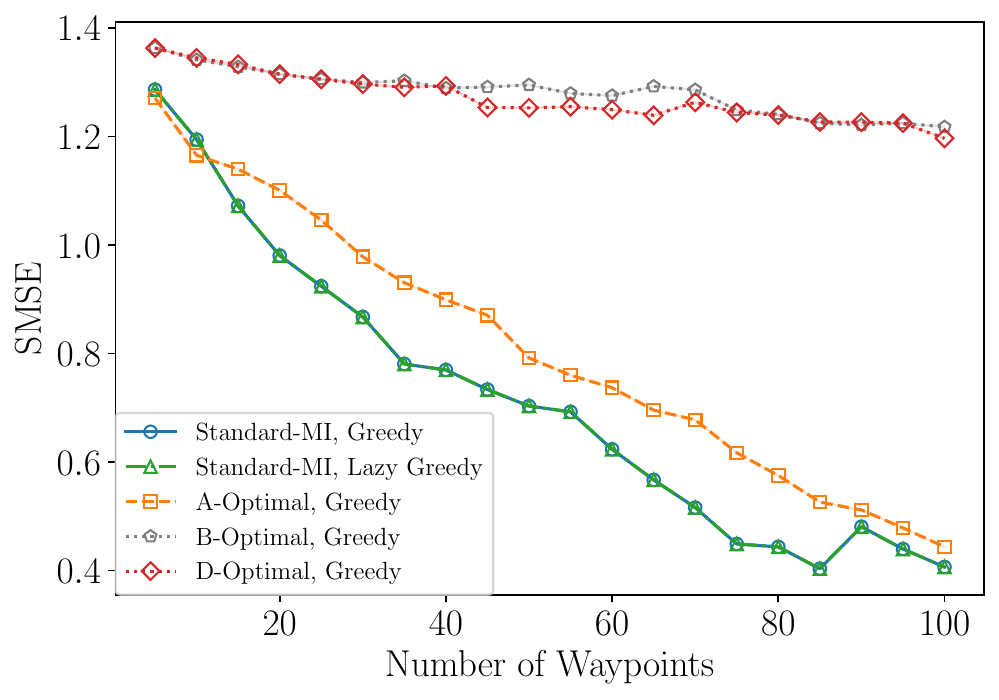}
        \caption{Mississippi}
    \end{subfigure}
    \hfill
    \begin{subfigure}{0.24\textwidth}
        \includegraphics[width=\textwidth]{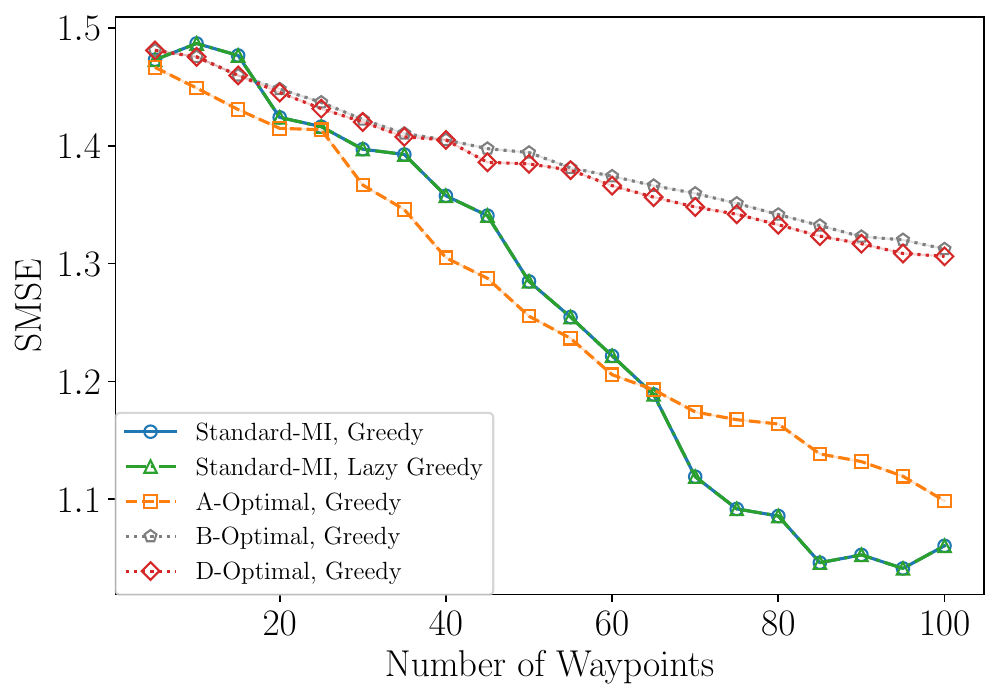}
        \caption{Nantucket}
    \end{subfigure}
    \hfill
    \begin{subfigure}{0.24\textwidth}
        \includegraphics[width=\textwidth]{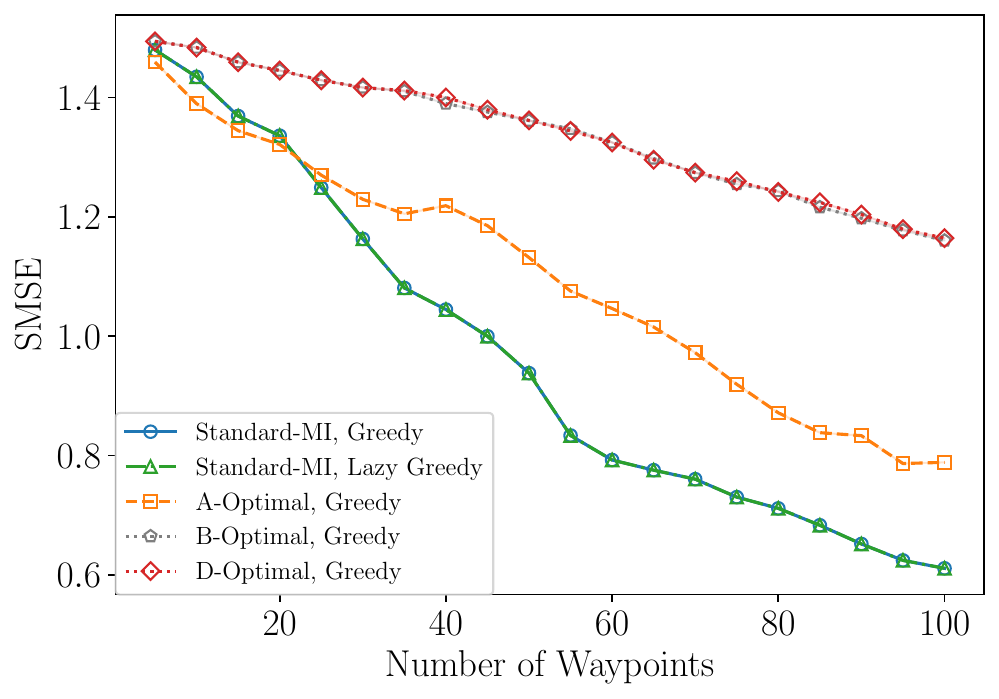}
        \caption{Virgin Islands}
    \end{subfigure}
    \hfill
    \begin{subfigure}{0.24\textwidth}
        \includegraphics[width=\textwidth]{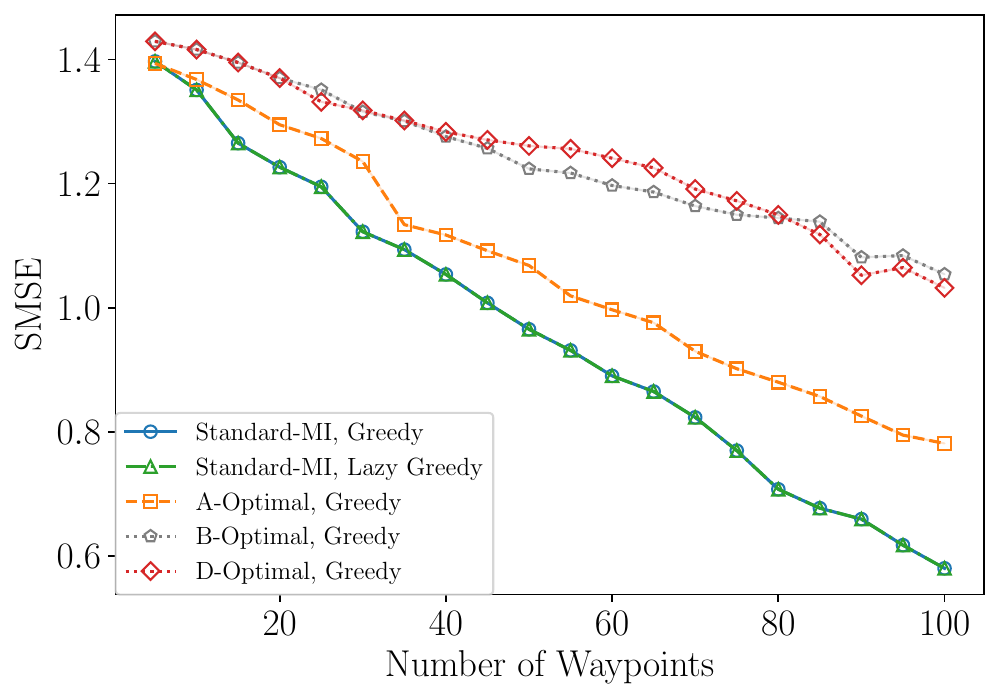}
        \caption{Wrangell}
    \end{subfigure}
    \caption{SMSE for sensor placement, comparing MI with Bayesian optimal design objectives. Curves show mean $\pm$ standard deviation of SMSE versus the number of waypoints; lower is better. Because the greedy methods are deterministic, results variability was negligible. These results highlight the performance advantage of MI-based objectives for discrete-space SP.}
    \label{fig:SMSE-sb1}
\end{figure*}

\begin{figure*}[!ht]
   \centering
    \begin{subfigure}{0.24\textwidth}
        \includegraphics[width=\textwidth]{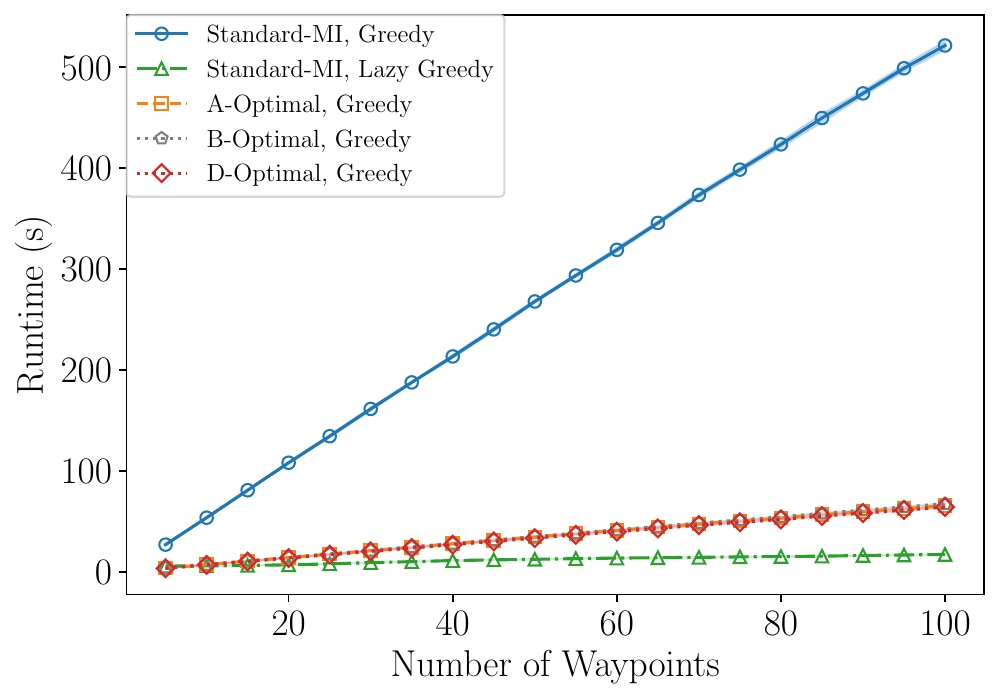}
        \caption{Mississippi}
    \end{subfigure}
    \hfill
    \begin{subfigure}{0.24\textwidth}
        \includegraphics[width=\textwidth]{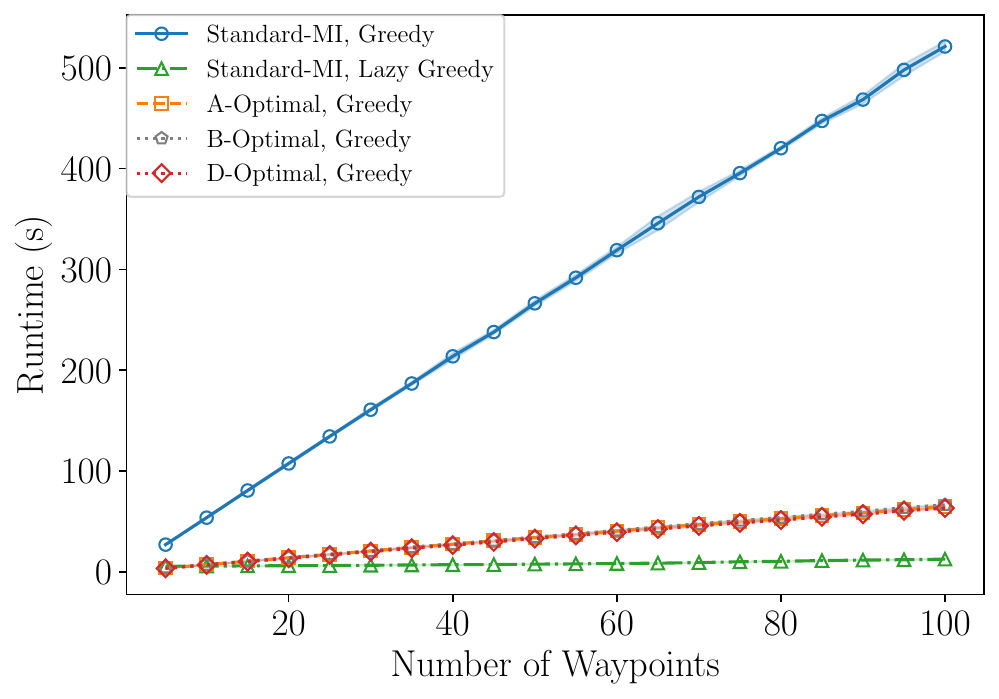}
        \caption{Nantucket}
    \end{subfigure}
    \hfill
    \begin{subfigure}{0.24\textwidth}
        \includegraphics[width=\textwidth]{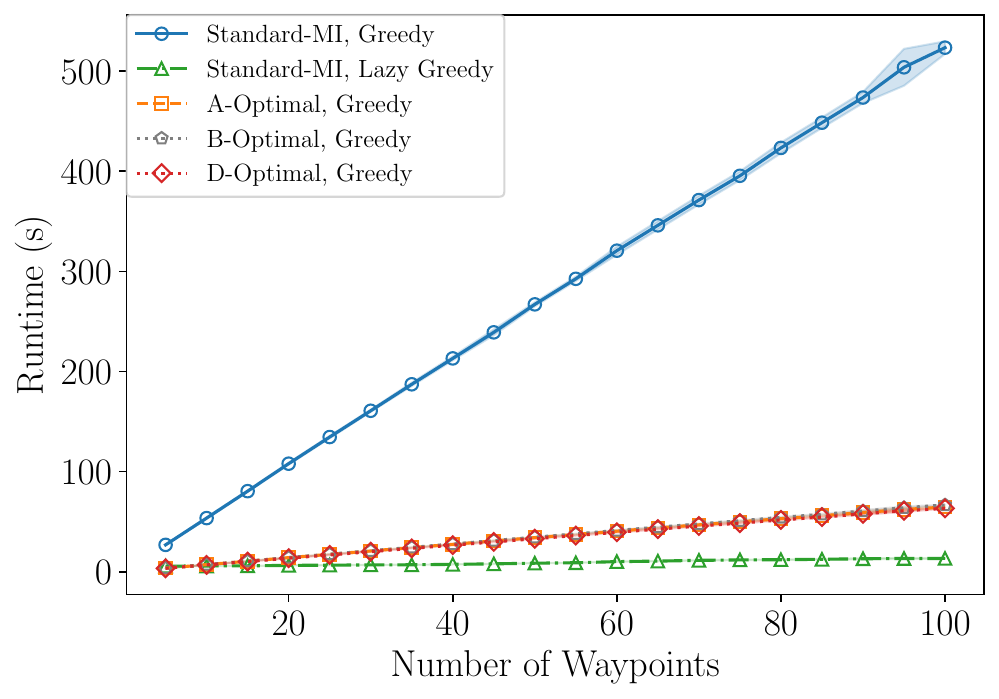}
        \caption{Virgin Islands}
    \end{subfigure}
    \hfill
    \begin{subfigure}{0.24\textwidth}
        \includegraphics[width=\textwidth]{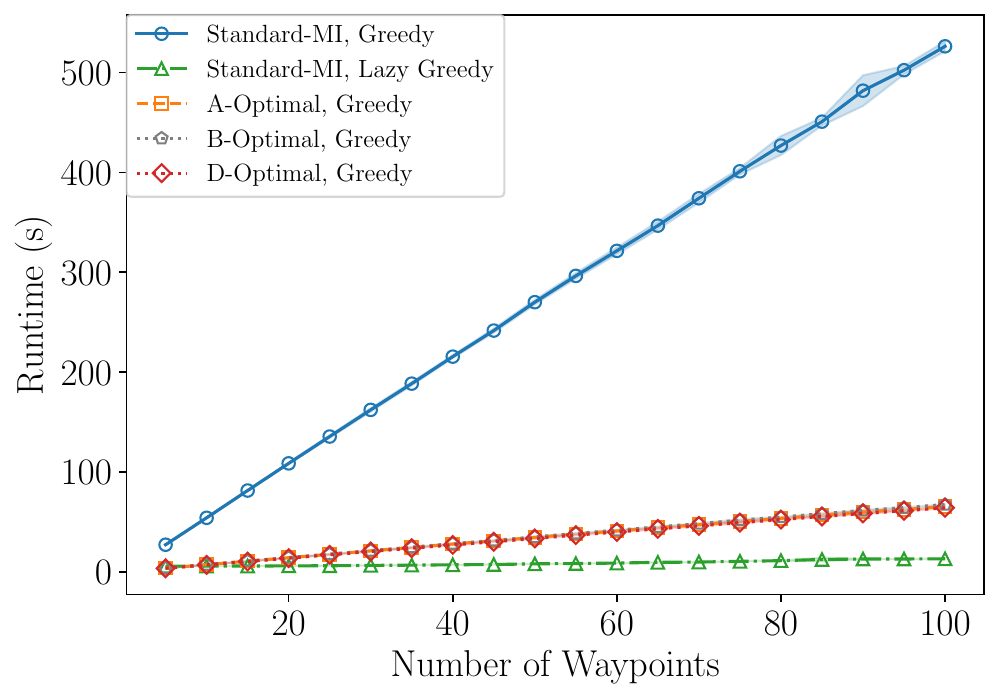}
        \caption{Wrangell}
    \end{subfigure}
    \caption{Runtime for sensor placement, comparing MI with Bayesian optimal design objectives. Curves show mean $\pm$ standard deviation of Runtime versus the number of waypoints; lower is better.}
    \label{fig:runtime-sb1}
\end{figure*}

\subsection{MI-Based Methods in Discrete Environments}
\label{exp:sb1}

We first benchmark Standard-MI against discrete-space optimal design metrics—A-, B-, and D-optimality~\cite{OttKB24}—to verify MI’s empirical advantages in sensor placement~(SP). All metrics were optimized using greedy selection. For MI, we additionally evaluated the lazy greedy algorithm~\cite{krauseSG08}, which retains the approximation guarantees of standard greedy while reducing the number of objective evaluations via a priority-queue strategy.

We evaluated SP on four bathymetry datasets—Mississippi, Nantucket, Virgin Islands, and Wrangell~\cite{datasets}. We used a Gaussian process~(GP) with a radial basis function~(RBF) kernel~\cite{RasmussenW05}, trained on $1000$ randomly sampled points from each environment, and a randomly sampled set of $500$ candidate sensing locations $\mathcal{V}$. Performance was evaluated for sensing set sizes from $5$ to $100$ in increments of $5$, with $10$ trials per setting.

Fig.~\ref{fig:SMSE-sb1} (standardized mean squared error; SMSE~\cite{RasmussenW05}) and Fig.~\ref{fig:runtime-sb1} (runtime) summarize the results. Consistent with prior work~\cite{krauseSG08}, MI outperforms the optimal-design baselines in reconstruction accuracy (SMSE). Greedy and lazy greedy achieve identical SMSE, but lazy greedy is substantially faster due to fewer MI evaluations. A-optimality attains unexpectedly strong SMSE on Nantucket, which we attribute to a mismatch between the stationary RBF kernel and the dataset’s non-stationary spatial structure. Overall, these results confirm that the Standard-MI objective optimized with lazy greedy remains a strong and reliable baseline for discrete-space sensor placement.

\begin{figure*}[!ht]
   \centering
    \begin{subfigure}{0.24\textwidth}
        \includegraphics[width=\textwidth]{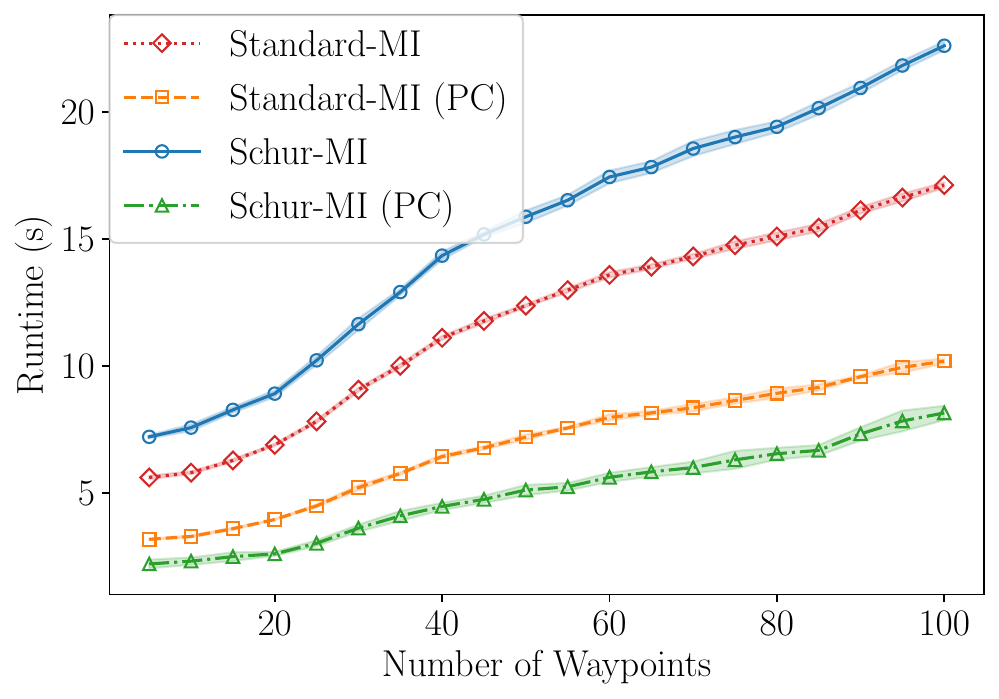}
        \caption{Mississippi}
    \end{subfigure}
    \hfill
    \begin{subfigure}{0.24\textwidth}
        \includegraphics[width=\textwidth]{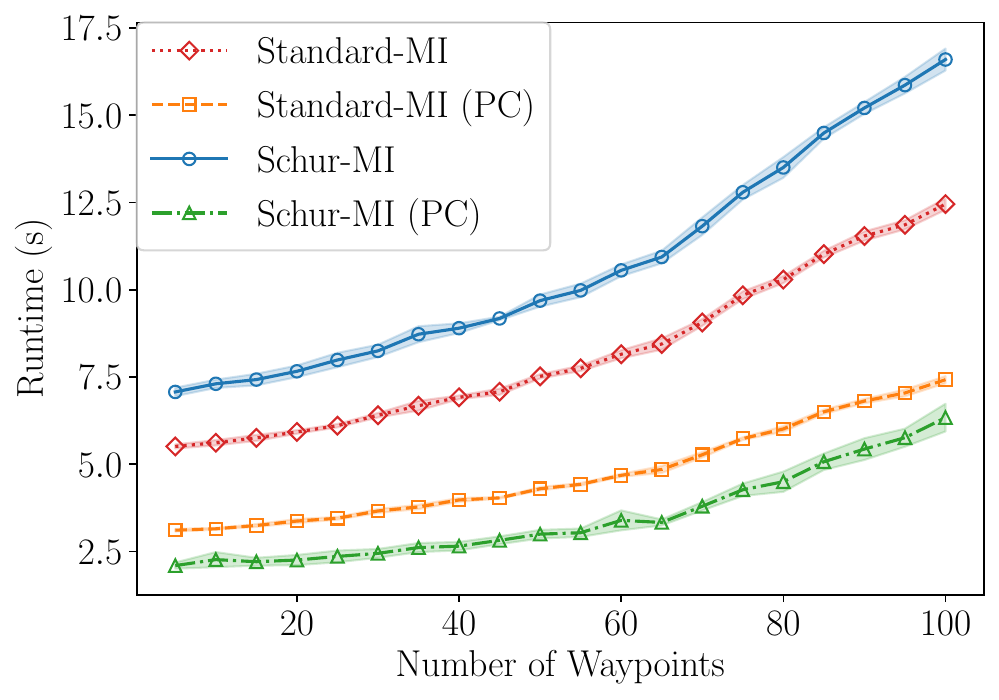}
        \caption{Nantucket}
    \end{subfigure}
    \hfill
    \begin{subfigure}{0.24\textwidth}
        \includegraphics[width=\textwidth]{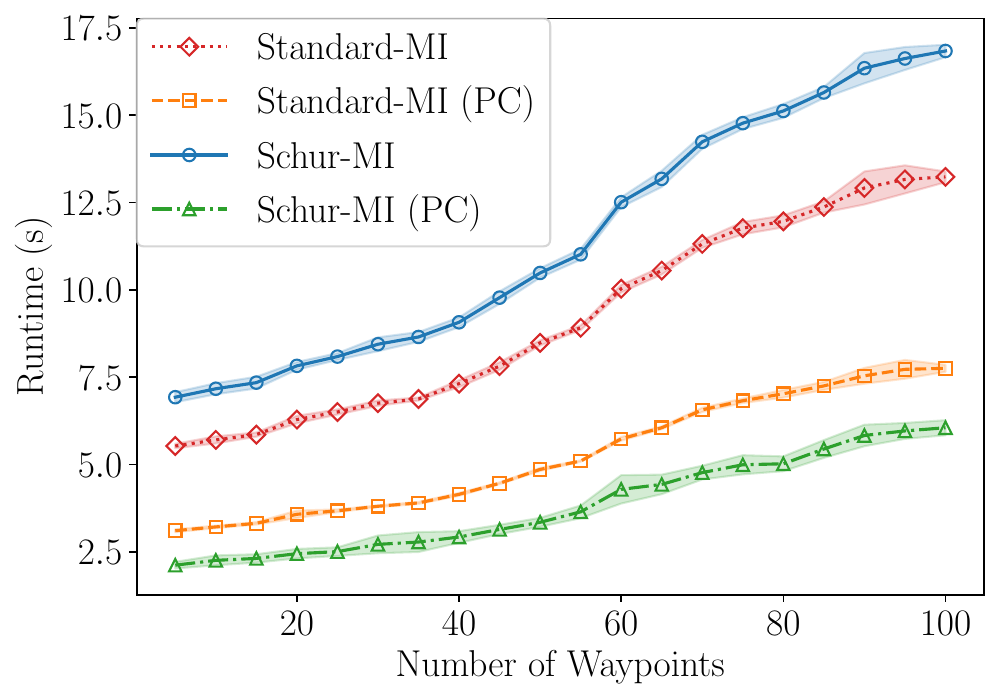}
        \caption{Virgin Islands}
    \end{subfigure}
    \hfill
    \begin{subfigure}{0.24\textwidth}
        \includegraphics[width=\textwidth]{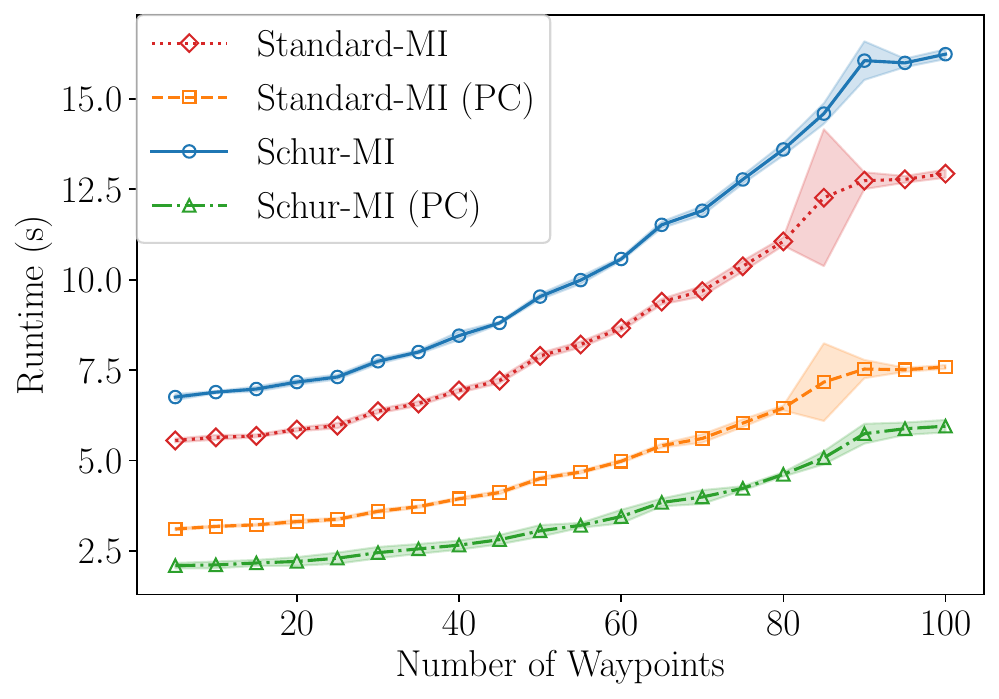}
        \caption{Wrangell}
    \end{subfigure}

    \caption{Runtime for SP across MI formulations. Curves show mean $\pm$ standard deviation of runtime versus the number of waypoints; lower is better. Schur-MI with PC is consistently the fastest MI formulation.}
    \label{fig:runtime-sb2}
\end{figure*}

\begin{figure*}[!ht]
   \centering
    \begin{subfigure}{0.24\textwidth}
        \includegraphics[width=\textwidth]{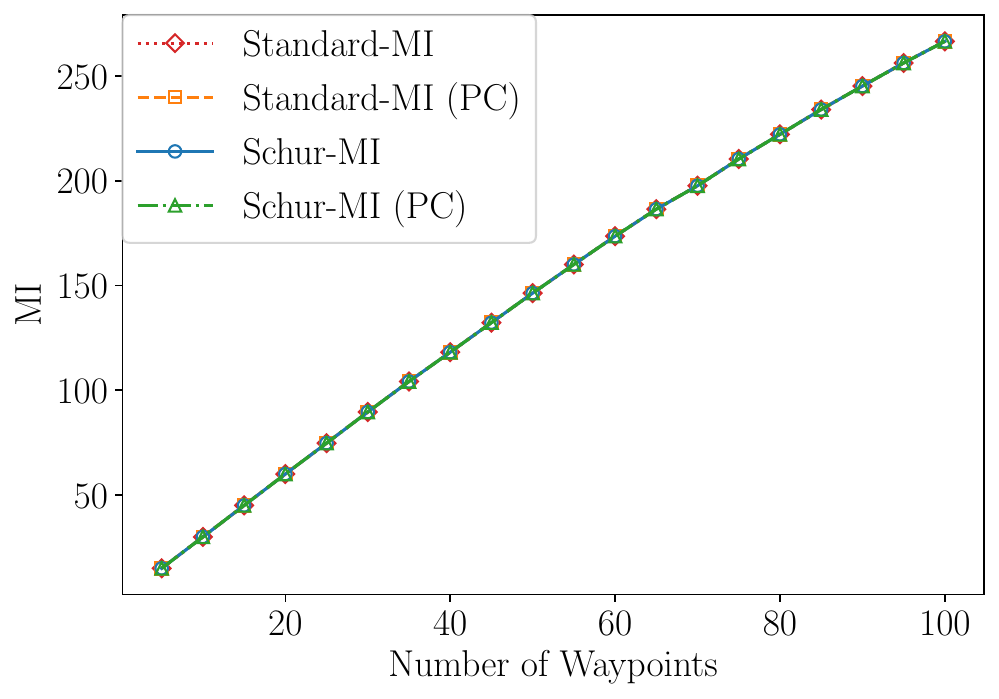}
        \caption{Mississippi}
    \end{subfigure}
    \hfill
    \begin{subfigure}{0.24\textwidth}
        \includegraphics[width=\textwidth]{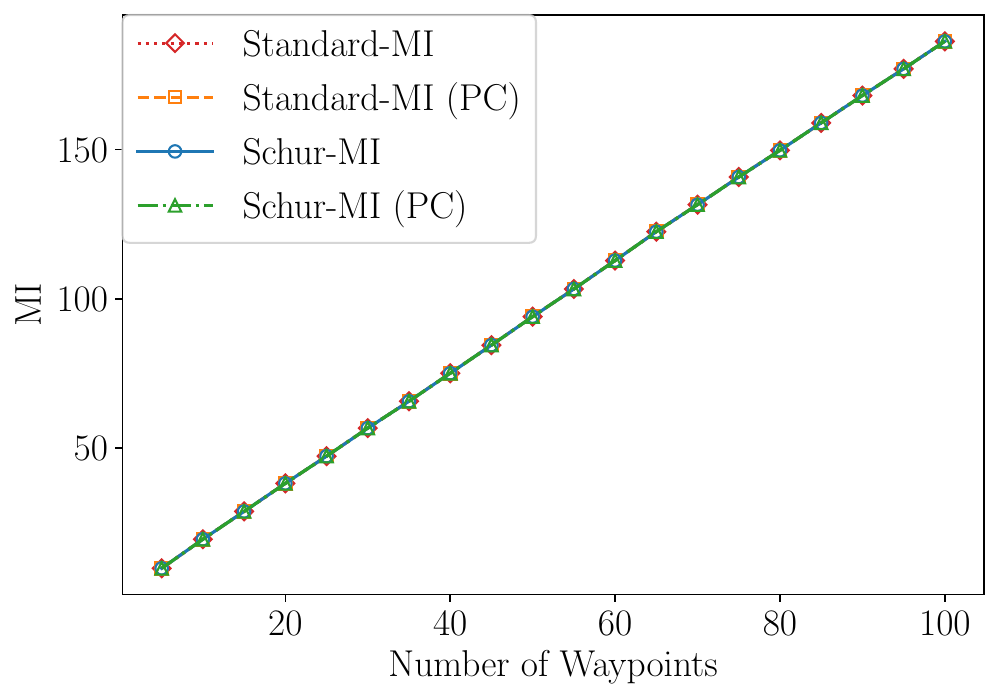}
        \caption{Nantucket}
    \end{subfigure}
    \hfill
    \begin{subfigure}{0.24\textwidth}
        \includegraphics[width=\textwidth]{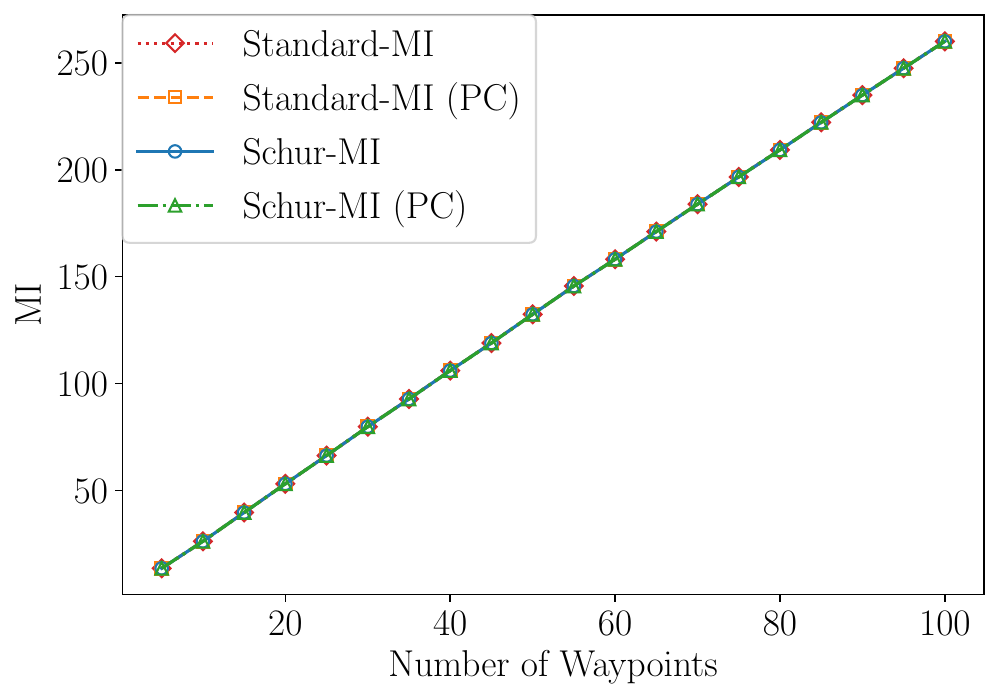}
        \caption{Virgin Islands}
    \end{subfigure}
    \hfill
    \begin{subfigure}{0.24\textwidth}
        \includegraphics[width=\textwidth]{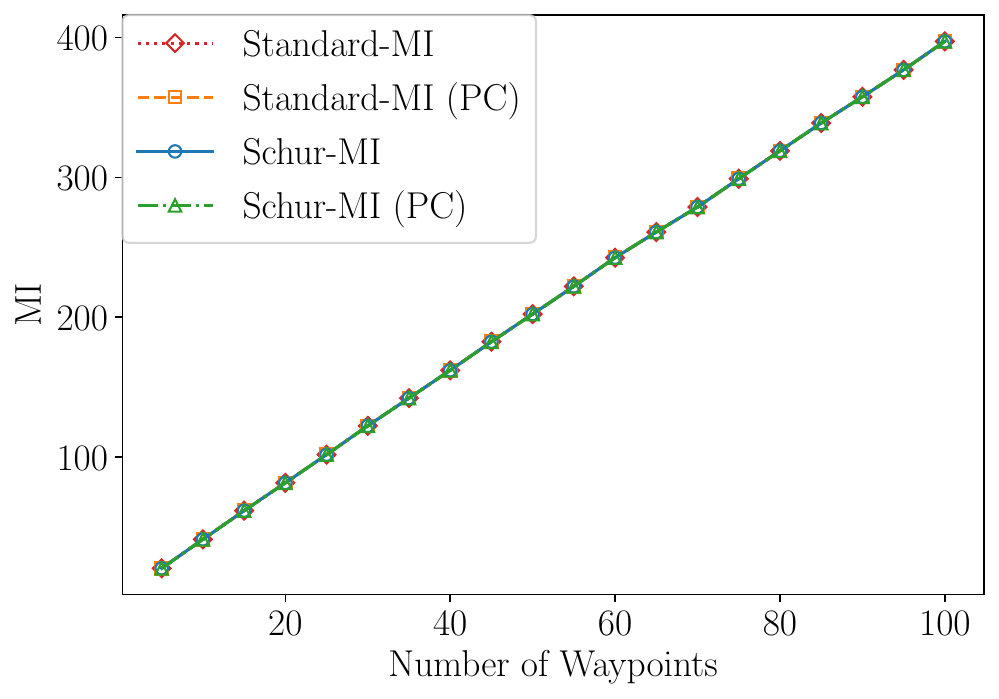}
        \caption{Wrangell}
    \end{subfigure}
    \caption{Standard-MI for SP across MI formulations. Curves show mean $\pm$ standard deviation of runtime versus the number of waypoints; lower is better. The results show that all four MI formulations are equivalent.}
    \label{fig:mi-sb2}
\end{figure*}

\begin{figure*}[!ht]
   \centering
    \begin{subfigure}{0.24\textwidth}
        \includegraphics[width=\textwidth]{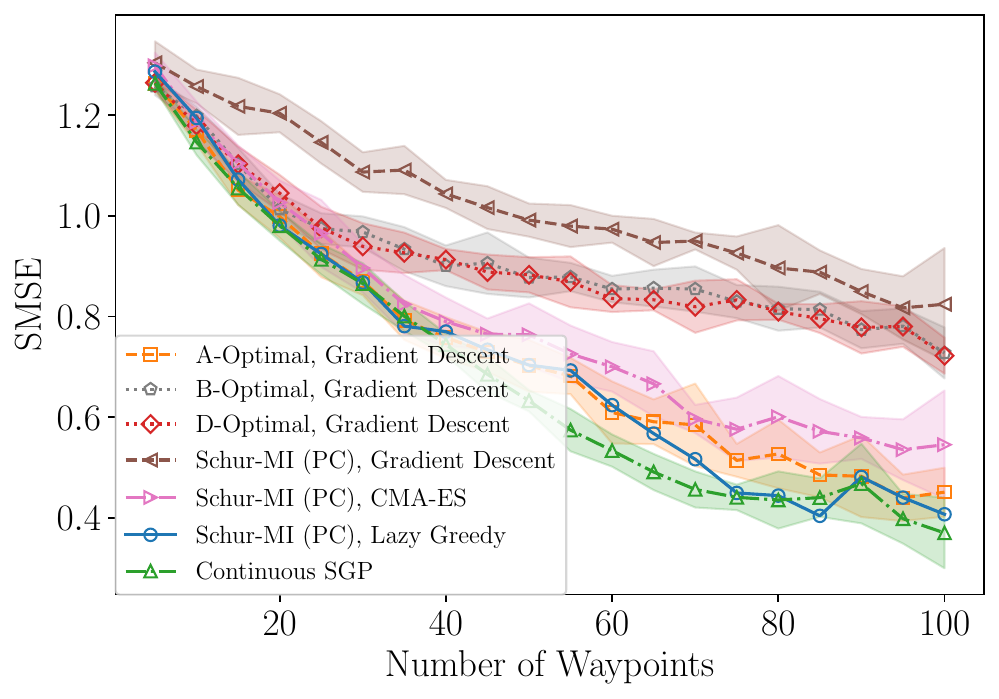}
        \caption{Mississippi}
    \end{subfigure}
    \hfill
    \begin{subfigure}{0.24\textwidth}
        \includegraphics[width=\textwidth]{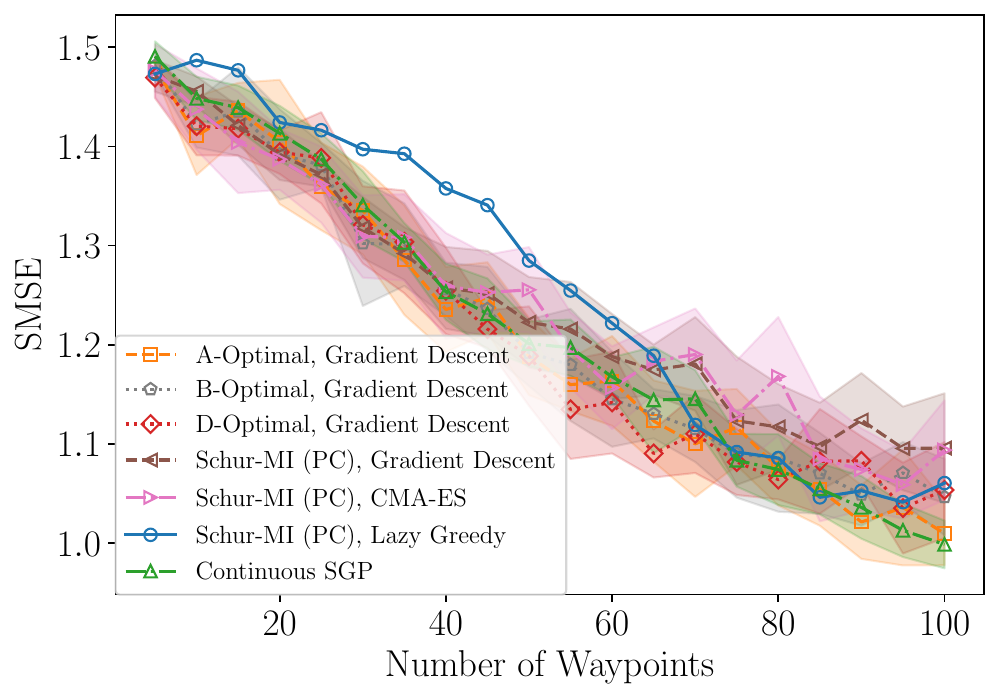}
        \caption{Nantucket}
    \end{subfigure}
    \hfill
    \begin{subfigure}{0.24\textwidth}
        \includegraphics[width=\textwidth]{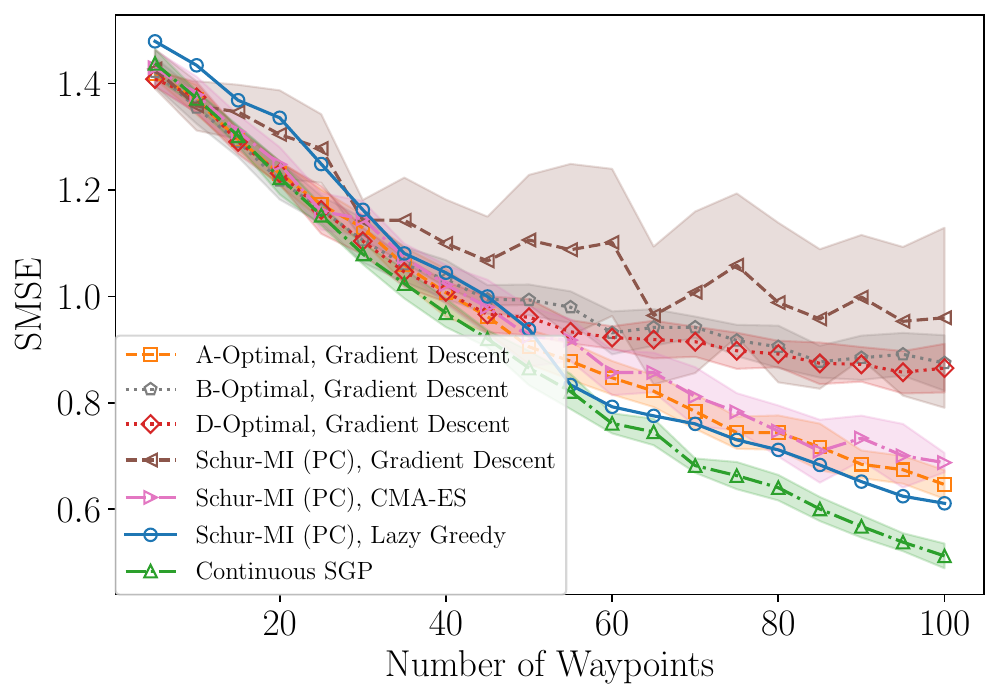}
        \caption{Virgin Islands}
    \end{subfigure}
    \hfill
    \begin{subfigure}{0.24\textwidth}
        \includegraphics[width=\textwidth]{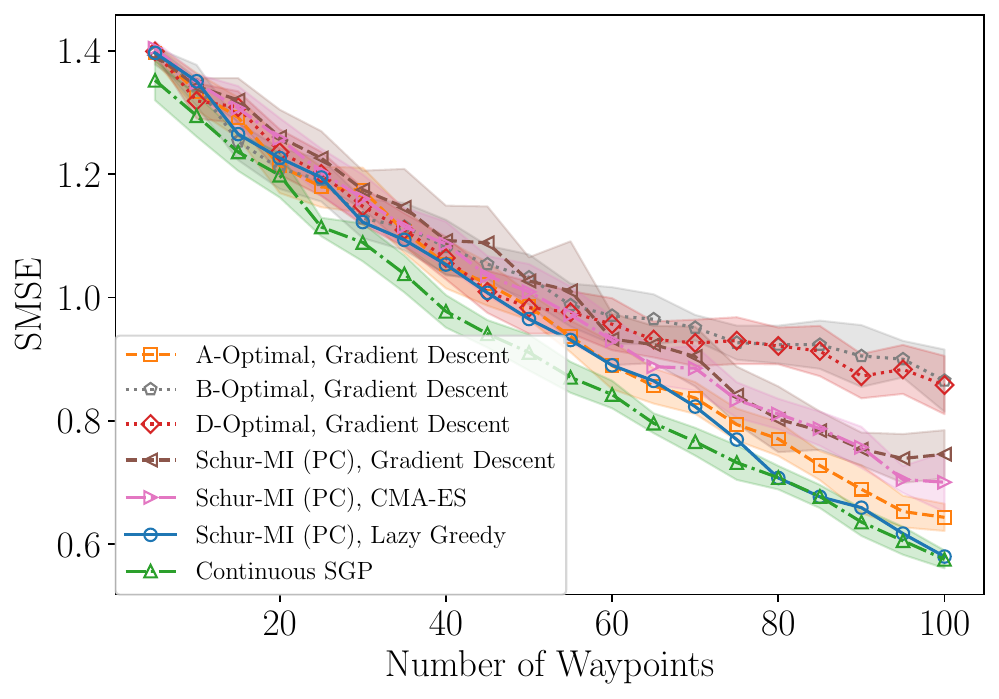}
        \caption{Wrangell}
    \end{subfigure}
    \caption{SMSE for SP in continuous space. Curves show mean $\pm$ standard deviation of SMSE versus the number of waypoints; lower is better. MI-based objectives remain competitive in continuous spaces.}
    \label{fig:SMSE-sb4}
\end{figure*}

\begin{figure*}[!ht]
   \centering
    \begin{subfigure}{0.24\textwidth}
        \includegraphics[width=\textwidth]{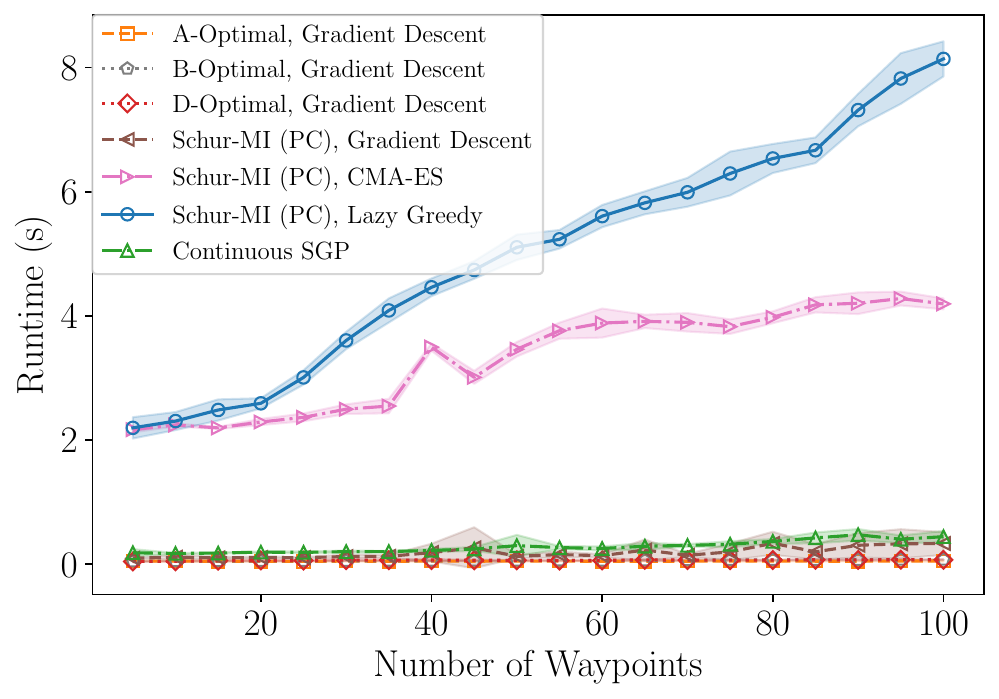}
        \caption{Mississippi}
    \end{subfigure}
    \hfill
    \begin{subfigure}{0.24\textwidth}
        \includegraphics[width=\textwidth]{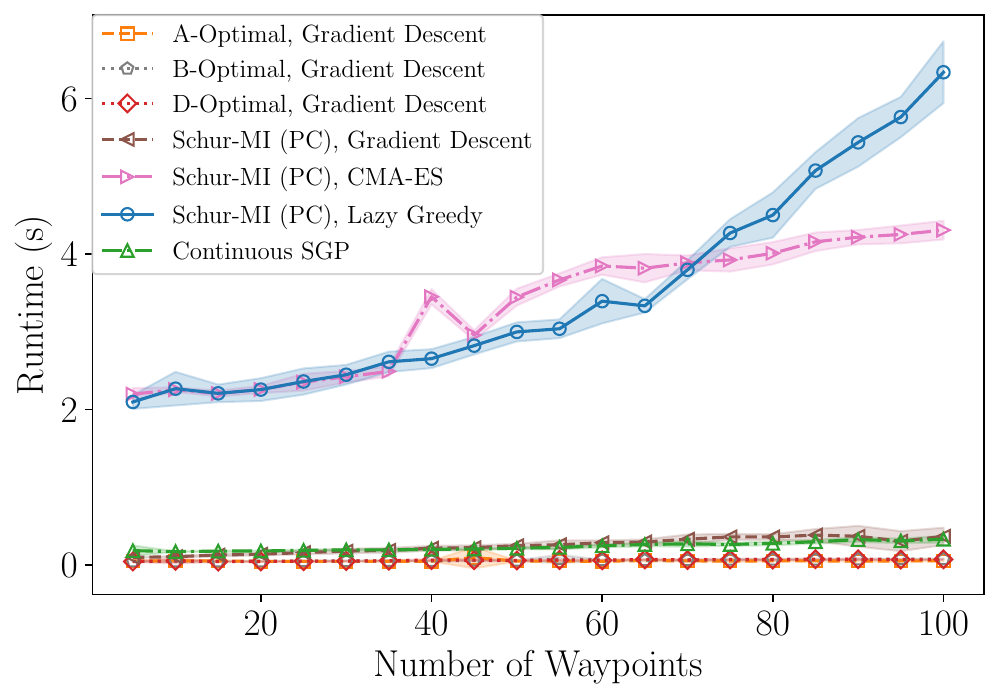}
        \caption{Nantucket}
    \end{subfigure}
    \hfill
    \begin{subfigure}{0.24\textwidth}
        \includegraphics[width=\textwidth]{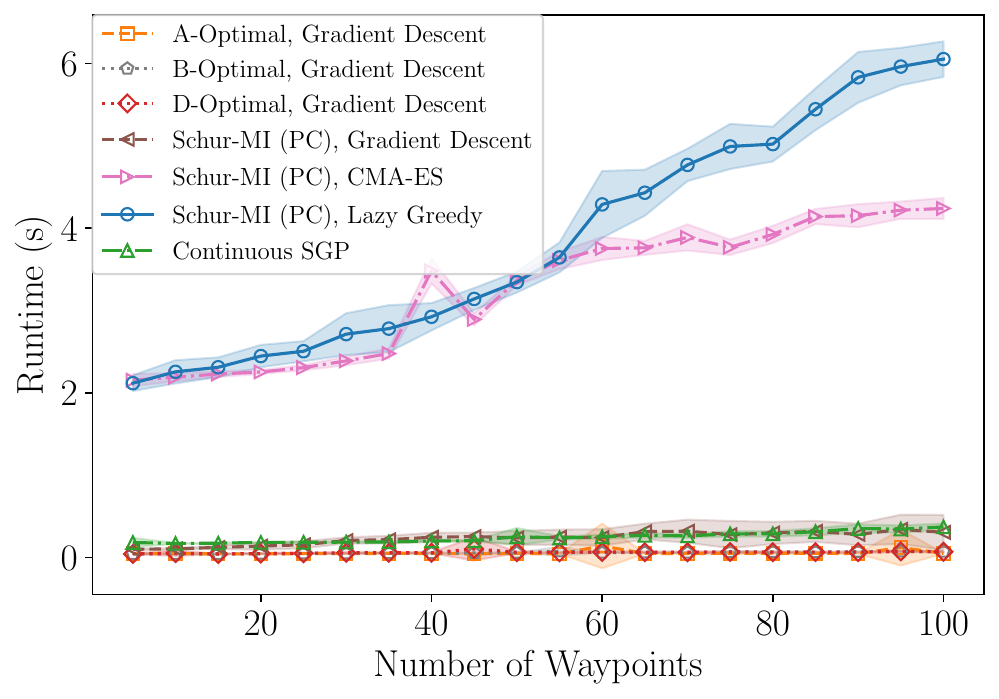}
        \caption{Virgin Islands}
    \end{subfigure}
    \hfill
    \begin{subfigure}{0.24\textwidth}
        \includegraphics[width=\textwidth]{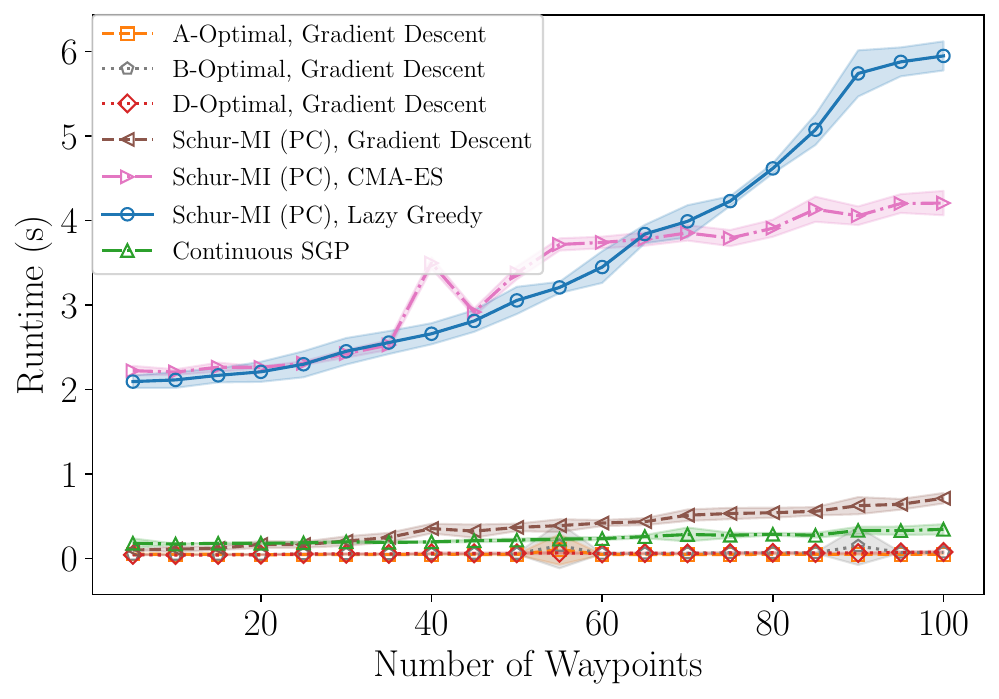}
        \caption{Wrangell}
    \end{subfigure}
    \caption{Runtime for SP in continuous space. Curves show mean $\pm$ standard deviation of Runtime versus the number of waypoints; lower is better. MI-based objectives remain competitive in continuous spaces.}
    \label{fig:runtime-sb4}
\end{figure*}

\subsection{Schur-MI with Precomputation}
\label{exp:sb2}

To empirically validate our formulations, we compare Schur-MI against Standard-MI, testing each with and without precomputation~(PC) under lazy greedy optimization. As detailed in Figures~\ref{fig:runtime-sb2} and~\ref{fig:mi-sb2}, which report the runtime and MI scores across the four datasets, all methods yield strictly identical results. This confirms the exact mathematical equivalence of the four formulations, entirely consistent with our theoretical derivations.

Without PC, Schur-MI is marginally slower than Standard-MI due to the constant-factor overhead of additional matrix multiplications inherent to the Schur formulation. However, when PC is enabled, Schur-MI emerges as the fastest approach, notably outperforming even Standard-MI with PC. This demonstrates the practical advantage of exposing reusable, invariant terms. Ultimately, these results highlight the critical role of our $\mathbf{I}(\mathcal{A}; \mathcal{V})$ formulation, which explicitly enables the efficient reuse of precomputed matrix factors across iterative MI evaluations.

\subsection{Continuous-Space Evaluation}
\label{exp:sb4}

We next evaluate SP in a continuous-space setting. Baselines include A-, B-, and D-optimality, each optimized via gradient descent. We compare against Schur-MI with PC optimized using (i) lazy greedy, (ii) gradient descent, and (iii) CMA-ES (a derivative-free evolutionary optimizer~\cite{HitzGGPS17}). We also include Continuous-SGP~\cite{JakkalaA24}, which leverages sparse GP approximations for scalability.

Fig.~\ref{fig:SMSE-sb4} and Fig.~\ref{fig:runtime-sb4} report SMSE and runtime, respectively. In reconstruction accuracy, Schur-MI (optimized via lazy greedy and CMA-ES), A-optimality, and Continuous-SGP perform similarly, achieving low SMSE across datasets. In contrast, Schur-MI optimized via gradient descent underperforms, likely due to sensitivity to local minima despite the objective being fully differentiable.

In runtime, A-optimality and Continuous-SGP are consistently the most efficient. Overall, these results suggest that for continuous-space sensor placement, A-optimality and Continuous-SGP currently provide the best reconstruction accuracy--runtime trade-off, while Schur-MI remains competitive, particularly when optimized with CMA-ES.

For discrete-space sensor placement, Schur-MI combined with PC yields the best overall performance, improving runtime relative to Standard-MI while preserving the near-optimal approximation guarantee. Together, these findings support Schur-MI as a principled and efficient objective for robotic information gathering.

\subsection{Schur-MI with Non-Stationary Kernels}
\label{exp:sb0}

\begin{figure}[!b]
   \centering
    \begin{subfigure}{0.49\linewidth}
        \includegraphics[width=\textwidth]{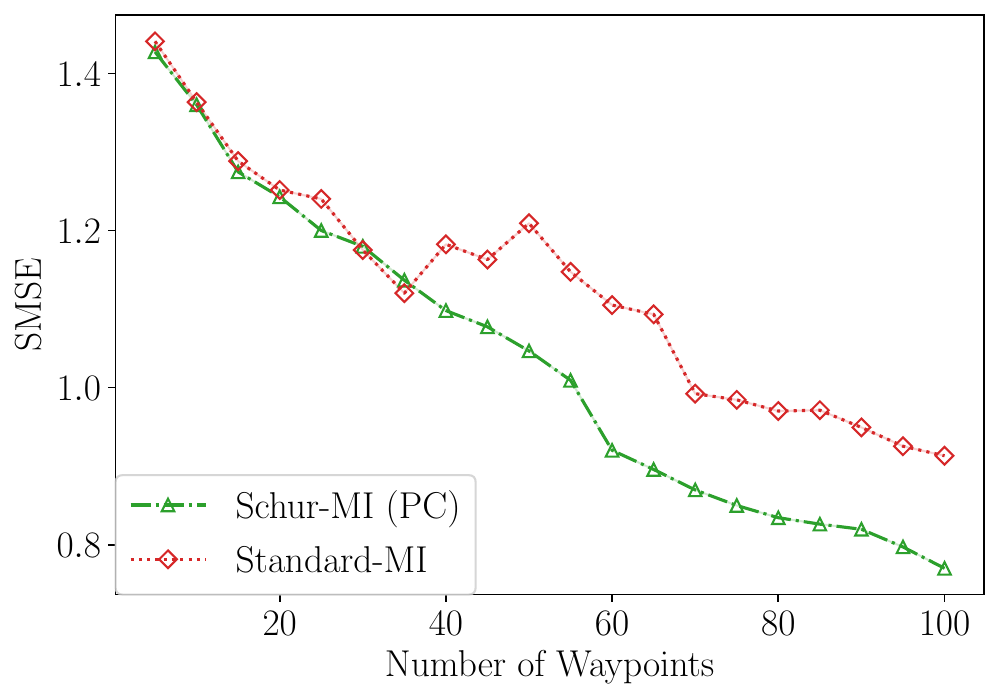}
        \caption{SMSE}
    \end{subfigure}
    \hfill
    \begin{subfigure}{0.49\linewidth}
        \includegraphics[width=\textwidth]{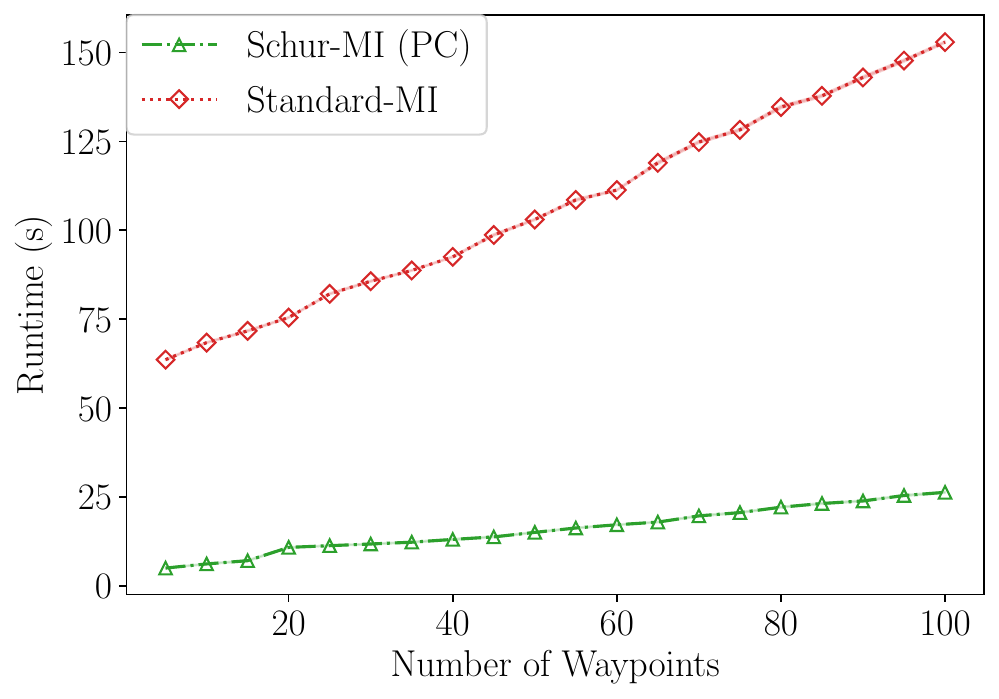}
        \caption{Runtime}
    \end{subfigure}
    \caption{SMSE (left) and runtime (right) for SP with a non-stationary kernel. Curves show mean $\pm$ standard deviation versus the number of waypoints; lower is better. Schur-MI achieves substantial gains with non-stationary kernels.}
    \label{fig:sb0}
\end{figure}

We further compare Schur-MI with PC to Standard-MI in a SP task using a GP with a non-stationary kernel~\cite{ChenKL22}. Unlike the stationary RBF kernel, the attentive kernel can capture spatially varying covariance structure. We use the Nantucket bathymetry dataset~\cite{datasets} and randomly sample 300 candidate sensing locations $\mathcal{V}$. SP was performed with lazy greedy~\cite{krauseSG08}. Fig.~\ref{fig:sb0} reports mean and standard deviation of SMSE, along with runtime. Given the deterministic nature of the optimization algorithm on a fixed candidate set, variance in both SMSE and runtime across repeated trials is negligible.

\begin{figure*}[!ht]
   \centering
    \includegraphics[width=\textwidth]{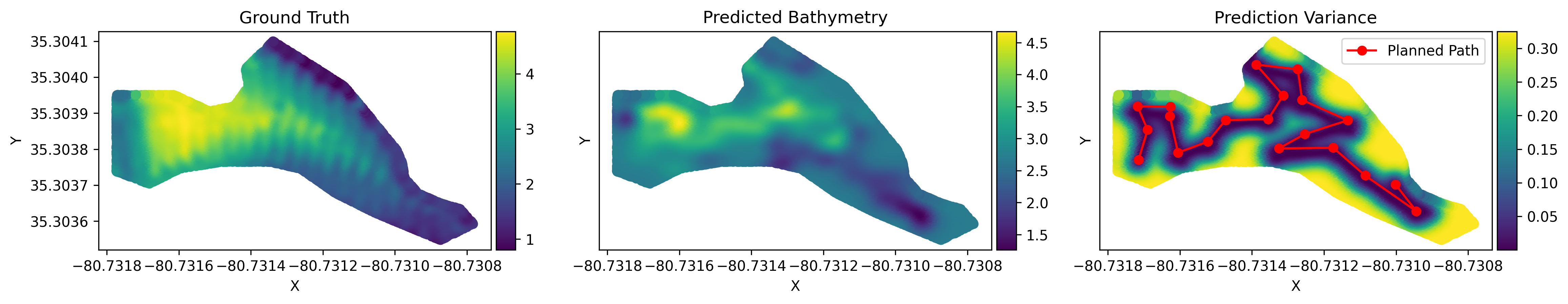}
    \caption{Field trial results comparing the ground truth (left) with the surface reconstruction obtained via the IPP solution data (middle), alongside the resulting predictive variance (right). All color bars indicate values in meters. The successful deployment highlights that Schur-MI is a theoretically rigorous and practical framework for RIG.}
    \label{fig:field_trial}
\end{figure*}

Schur-MI consistently achieves a lower SMSE than Standard-MI, with the performance gap widening for larger sensing sets. We attribute this to the original RIG formulation, $\mathbb{I}(\mathcal{A};\mathcal{V}\setminus\mathcal{A})$, used in Standard-MI, which can induce a spatial bias toward unsensed regions. In contrast, Schur-MI employs the $\mathbb{I}(\mathcal{A};\mathcal{V})$ RIG formulation; this eliminates the bias by measuring information gain with respect to the environment as a whole. Additionally, Schur-MI with PC yields up to a \emph{12.7$\times$ speedup} over the standard formulation. These results emphasize both the accuracy and computational advantages of Schur-MI with PC, particularly under non-stationary modeling.

\subsection{Field Trial: IPP for Bathymetry Mapping using an ASV}
\label{exp:field_trial}

To assess real-world applicability, we deployed our method on an autonomous surface vehicle~(ASV)~\cite{Brancato22} in a lake. The ASV~(Fig.~\ref{fig:cover}) was equipped with GPS, a single-beam sonar, and a Raspberry Pi 5 for onboard computation.

We conducted an informative path planning~(IPP) trial in which the planner iteratively updated an online GP model by updating hyperparameters from collected data and replanning the remaining waypoints to maximize information gain. Following the IPP framework of Jakkala and Akella~\cite{JakkalaA25}, we used Schur-MI with PC as the objective, an RBF kernel for the GP model, and optimized 20 waypoints using CMA-ES.

To evaluate reconstruction quality, we performed a dense boustrophedon survey of the environment to serve as ground truth. The IPP-generated data was then used to reconstruct the bathymetry and the posterior variance (Fig.~\ref{fig:field_trial}). The reconstruction achieved an SMSE of 4.54 and an RMSE of 0.60. The disparity between these metrics is attributed to elevated predictive uncertainty in unvisited regions—an expected outcome given the RBF kernel’s finite length-scale.

Overall, the ASV successfully executed an adaptive, data-driven path under real-time sensing and compute constraints. These results demonstrate that Schur-MI is theoretically well-grounded and also practical for field-deployable RIG.
\section{Conclusion}

This paper presented Schur-MI, a computationally efficient mutual information~(MI) objective for Gaussian process (GP)–based robotic information gathering~(RIG). By combining a Schur-complement factorization of the MI criterion with a precomputation strategy tailored to the iterative nature of RIG planners, Schur-MI dramatically reduces the per-evaluation cost of MI from $\mathcal{O}(|\mathcal{V}|^3)$ to $\mathcal{O}(|\mathcal{A}|^3)$, where $\mathcal{V}$ and $\mathcal{A}$ represent the candidate and selected sensing sets, respectively.

Extensive evaluations across discrete and continuous domains demonstrate that Schur-MI preserves the rigorous informativeness of Standard-MI while delivering substantial runtime improvements. In continuous spaces, it performs competitively alongside established baselines such as A-optimal design and Continuous-SGP. In discrete settings, it consistently outperforms Standard-MI in computational speed and achieves comparable or superior reconstruction accuracy, all while maintaining the near-optimal approximation guarantees of greedy MI-based selection. Furthermore, field trials with a compute-constrained autonomous surface vehicle~(ASV) validate Schur-MI's practical efficacy for real-world, adaptive informative path planning~(IPP).

Future work will rigorously analyze the surrogate-set formulation, focusing on its sensitivity to perturbation magnitudes and the resulting impact on solution quality. Ultimately, by rendering MI evaluation tractable for iterative online planning, \textbf{Schur-MI successfully bridges the longstanding gap between principled information-theoretic objectives and real-time robotic exploration}.

\bibliography{references}


\end{document}